\documentclass[a4paper,fleqn]{cas-dc}
\usepackage[linesnumbered,ruled]{algorithm2e}
\usepackage{times}
\usepackage{soul}
\usepackage{url}
\usepackage[small]{caption}
\usepackage{graphicx}
\usepackage{amsmath}
\usepackage{amssymb}
\usepackage{amsthm}
\usepackage{colortbl,booktabs}
\usepackage{threeparttable}
\usepackage{multirow}
\usepackage{array}
\usepackage{bbding}
\usepackage{stfloats}
\usepackage[numbers]{natbib}
\usepackage[switch]{lineno}
\usepackage{lettrine}
\usepackage{bm}
\usepackage{hyperref}

\def\tsc#1{\csdef{#1}{\textsc{\lowercase{#1}}\xspace}}
\tsc{WGM}
\tsc{QE}
\tsc{EP}
\tsc{PMS}
\tsc{BEC}
\tsc{DE}

\begin{document}

\let\WriteBookmarks\relax
\def\floatpagepagefraction{1}
\def\textpagefraction{.001}

\shorttitle{Neurocomputing 500 (2022) 832--845}

\shortauthors{https://doi.org/10.1016/j.neucom.2022.05.091}

\title[mode=title]{Deep Multi-View Semi-Supervised Clustering with Sample Pairwise Constraints}

\author[1,2,3]{Rui Chen}
\ead{chenrui_1216@163.com}
\author[2]{Yongqiang Tang}\cormark[1]
\ead{yongqiang.tang@ia.ac.cn}
\author[1,2]{Wensheng Zhang}\cormark[1]
\ead{zhangwenshengia@hotmail.com}
\author[1,3]{Wenlong Feng}
\ead{fwlfwl@163.com}

\affiliation[1]{organization={College of Information Science and Technology, Hainan University}, city={Haikou}, postcode={570208}, country={China}}
\affiliation[2]{organization={Research Center of Precision Sensing and Control, Institute of Automation, Chinese Academy of Sciences}, city={Beijing}, postcode={100190}, country={China}}
\affiliation[3]{organization={State Key Laboratory of Marine Resource Utilization in South China Sea, Hainan University}, city={Haikou}, postcode={570208}, country={China}}

\cortext[cor1]{Corresponding author}


\begin{abstract}
Multi-view clustering has attracted much attention thanks to the capacity of multi-source information integration. Although numerous advanced methods have been proposed in past decades, most of them generally overlook the significance of weakly-supervised information and fail to preserve the feature properties of multiple views, thus resulting in unsatisfactory clustering performance. To address these issues, in this paper, we propose a novel \textbf{D}eep \textbf{M}ulti-view \textbf{S}emi-supervised \textbf{C}lustering (\textbf{DMSC}) method, which jointly optimizes three kinds of losses during networks finetuning, including multi-view clustering loss, semi-supervised pairwise constraint loss and multiple autoencoders reconstruction loss. Specifically, a KL divergence based multi-view clustering loss is imposed on the common representation of multi-view data to perform heterogeneous feature optimization, multi-view weighting and clustering prediction simultaneously. Then, we innovatively propose to integrate pairwise constraints into the process of multi-view clustering by enforcing the learned multi-view representation of must-link samples (cannot-link samples) to be similar (dissimilar), such that the formed clustering architecture can be more credible. Moreover, unlike existing rivals that only preserve the encoders for each heterogeneous branch during networks finetuning, we further propose to tune the intact autoencoders frame that contains both encoders and decoders. In this way, the issue of serious corruption of view-specific and view-shared feature space could be alleviated, making the whole training procedure more stable. Through comprehensive experiments on eight popular image datasets, we demonstrate that our proposed approach performs better than the state-of-the-art multi-view and single-view competitors.
\end{abstract}


\maketitle

\section{Introduction}\label{section:Introduction}
\lettrine{\textbf{C}}{lustering}, a crucial but challenging topic in both data mining and machine learning communities, aims to partition the data into different groups such that samples in the same group are more similar to each other than to those from other groups. Over the past few decades, various efforts have been exploited, such as prototype-based clustering \cite{KM, DEC}, graph-based clustering \cite{SC, SDCN}, model-based clustering \cite{GMM, DGG}, density-based clustering \cite{DBSCAN, DDC}, etc. With the prevalence of deep learning technology, many researches have integrated the powerful nonlinear embedding capability of deep neural networks (DNNs) into clustering, and achieved dazzling clustering performance. Xie et al. \cite{DEC} make use of DNNs to mine the cluster-oriented feature for raw data, realizing a substantial improvement compared with conventional clustering techniques. Bo et al. \cite{SDCN} combine autoencoder representation with graph embedding and propose a structural deep clustering network (SDCN) owning a better performance over the other baseline methods. Yang et al. \cite{DGG} develop a variational deep Gaussian mixture model (GMM) \cite{GMM} to facilitate clustering. Ren et al. \cite{DDC} present a deep density-based clustering (DDC) approach, which is able to adaptively estimate the number of clusters with arbitrary shapes. Despite the great success of deep clustering methods, they can only be satisfied with single-view clustering scenarios.

In the real-world applications, data are usually described as various heterogeneous views or modalities, which are mainly collected from multiple sensors or feature extractors. For instance, in computer vision, images can be represented by different hand-crafted visual features such as Gabor \cite{Gabor}, LBP \cite{LBP}, SIFT \cite{SIFT}, HOG \cite{HOG}; in information retrieval, web pages can be exhibited by page text or links to them; in intelligent security, one person can be identified by face, fingerprint, iris, signature; in medical image analysis, a subject may have a binding relationship with different types of medical images (e.g., X-ray, CT, MRI). Obviously, single-view based methods are no longer suitable for such multi-view data, and how to cluster this kind of data is still a long-standing challenge on account of the inefficient incorporation of multiple views. Consequently, numerous multi-view clustering applications have been developed to jointly deal with several types of features or descriptors.

Canonical correlation analysis (CCA) \cite{CCA} seeks two projections to map two views onto a low-dimensional common subspace, in which the linear correlation between the two views is maximized. Kernel canonical correlation analysis (KCCA) \cite{KCCA} resolves more complicated correlations by equipping the kernel trick into CCA. Multi-view subspace clustering methods \cite{MVSC, LMSC, CoMSC, CTRL, JSTC, T-MEK-SPL} are aimed at utilizing multi-view data to reveal the potential clustering architecture, most of which usually devise multi-view regularizer to describe the inter-view relationships between different formats of features. In recent years, a variety of DNNs-based multi-view learning algorithms have emerged one after another. Deep canonical correlation analysis (DCCA) \cite{DCCA} and deep canonically correlated autoencoders (DCCAE) \cite{DCCAE} successfully draw on DNNs’ advantage of nonlinear mapping and improve the representation capacity of CCA. Deep generalized canonical correlation analysis (DGCCA) \cite{DGCCA} combines the effectiveness of deep representation learning with the generalization of integrating information from more than two independent views. Deep embedded multi-view clustering (DEMVC) \cite{DEMVC} learns the consistent and complementary information from multiple views with a collaborative training mechanism to heighten clustering effectiveness. Autoencoder in autoencoder network (AE$^{2}$-Net) \cite{AE2-Nets} jointly learns view-specific feature for each view and encodes them into a complete latent representation with a deep nested autoencoder framework. Cognitive deep incomplete multi-view clustering network (CDIMC-net) \cite{CDIMC-net} incorporates DNNs pretraining, graph embedding and self-paced learning to enhance the robustness of marginal samples while maintaining the local structure of data, and a superior performance is accomplished.

Despite these excellent achievements, current deep multi-view clustering methods still present two obvious drawbacks. Firstly, most previous approaches fail to take advantage of semi-supervised prior knowledge to guide multi-view clustering. It is known that pairwise constraints are easy to obtain in practice and have been frequently utilized in many semi-supervised learning scenes \cite{semi-KM 1, semi-KM 2, semi-SC}. Therefore, ignoring this kind of precious weakly-supervised information will undoubtedly place restrictions on the model performance. Meanwhile, the constructed clustering structure is likely to be unreasonable and imperfect as well. Besides, one more issue attracting our attention is that most existing studies typically cast away the decoding networks during the finetuning process while overlooking the preservation of feature properties. Such an operation may cause serious corruption of both view-specific and view-shared feature space, thus hindering the clustering performance accordingly.

In order to settle the aforementioned defectiveness, we propose a novel Deep Multi-view Semi-supervised Clustering (DMSC) method in this paper. Our method embodies two stages: 1) parameters initialization, 2) networks finetuning. In the initialization stage, we pretrain multiple deep autoencoder branches by minimizing their reconstruction losses end-to-end to extract high-level compact feature for each view. In the finetuning stage, we consider three loss items, i.e., multi-view clustering loss, semi-supervised pairwise constraint loss and multiple autoencoders reconstruction loss. Specifically, for multi-view clustering loss, we adopt the KL divergence based soft assignment distribution strategy proposed by the pioneering work \cite{DMJCS} to perform heterogeneous feature optimization, multi-view weighting and clustering prediction simultaneously. Then, in order to exploit the weakly-supervised pairwise constraint information that plays a key role in shaping a reasonable latent clustering structure, we introduce a constraint matrix and enforce the learned multi-view common representation to be similar for must-link samples and dissimilar for cannot-link samples. For multiple autoencoders reconstruction loss, we tune the intact autoencoder frame for each heterogeneous branch, such that view-specific attributes can be well protected to evade the unexpected destruction of the corresponding feature domain. Through this way, our learned conjoint representation could be more robust than that in rivals who only hold back the encoder part during finetuning. To sum up, the main contributions of this work are highlighted as follows:
\begin{itemize}
	\item We innovatively propose a deep multi-view semi-supervised clustering approach termed DMSC, which can utilize the user-given pairwise constraints as weak supervision to lead cluster-oriented representation learning for joint multi-view clustering.
	\item During networks finetuning, we introduce the feature structure preservation mechanism into our model, which is conducive to ensuring both distinctiveness of the local specific view and completeness of the global shared view.
	\item The proposed DMSC enjoys the strength of efficiently digging out the complementary information hidden in different views and the cluster-friendly discriminative embeddings to rouse model performance.
	\item Comprehensive comparison experiments on eight widely used benchmark image datasets demonstrate that our DMSC possesses superior clustering performance against the state-of-the-art multi-view and single-view competitors. The elaborate experimental analysis confirms the effectiveness and generalization of the proposed approach.
\end{itemize}

The remainder of this paper is organized as follows. In Section \ref{section:Related Work}, we make a brief review on the related work. Section \ref{section:Methodology} describes the details of the developed DMSC algorithm. Extensive experimental results are reported and analyzed in Section \ref{section:Experiment}. Finally, Section \ref{section:Conclusion} concludes this paper.

\section{Related Work}\label{section:Related Work}
This section reviews some of the previous researches closely related to this paper. We first briefly review a few antecedent works on deep clustering. Then, related studies of multi-view clustering are reviewed. Finally, we introduce the semi-supervised clustering paradigm.

\subsection{Deep Clustering}
Existing deep clustering approaches can be generally partitioned into two categories. One category covers methods that usually treat representation learning and clustering separately, i.e., project the original data into a low-dimensional feature space first, and then perform traditional clustering algorithms \cite{KM, SC, GMM, DBSCAN} to group feature points. Unfortunately, this kind of independent form may restrict the clustering performance due to the oversight of some underlying relationships between representation learning and clustering. Another category refers to methods that apply the joint optimization criterion, which perform both representation learning and clustering simultaneously, showing considerable superiority beyond the separated counterparts. Recently, several attempts have been proposed to integrate representation learning and clustering into a unified framework. Inspired by $t$-SNE \cite{t-SNE}, Xie et al. \cite{DEC} propose a deep embedded clustering (DEC) model to utilize a stacked autoencoder (SAE) to excavate the high-level representation for input data, then iteratively optimize a KL divergence based clustering objective with the help of auxiliary target distribution. Guo et al. \cite{IDEC} further put forward to integrate SAE’s reconstruction loss into the DEC objective to avoid corrosion of the embedded space, bringing about appreciable advancement. Yang et al. \cite{DCN} combine SAE-based cluster-oriented dimensionality reduction and $K$-means \cite{KM} clustering together to jointly enhance the performance of both, which requires an alternative optimization strategy to discretely update cluster centers, cluster pseudo labels and network parameters. Drawing on the experience of hard-weighted self-paced learning, Guo et al. \cite{ASPC} and Chen et al. \cite{DCSPC} prioritize high-confidence samples during the clustering network training to buffer the negative impact of outliers and steady the whole training process. Ren et al. \cite{SDEC} overcome the vulnerability in DEC that fails to guide the clustering by making use of prior information. Li et al. \cite{DBC} present a discriminatively boosted clustering framework with the help of a convolutional feature extractor and a soft assignment model. Fard et al. \cite{DKM} raise an approach for jointly clustering by reconsidering the $K$-means loss as the limit of a differentiable function that touches off a truly solution.

\subsection{Multi-View Clustering}
Multi-view clustering \cite{review, SAMVC, DCMSC, MCIM} aims to utilize the available multi-view features to learn common representation and perform clustering to obtain data partitions. With regard to shallow methods, Cai et al. \cite{RMKMC} propose a robust multi-view $K$-means clustering (RMKMC) algorithm by introducing a shared indicator matrix across different views. Xu et al. \cite{MSPL} develop an improved version of RMKMC to learn the multi-view model by simultaneously considering the complexities of both samples and views, relieving the local minima problem. Zhang et al. \cite{BMVC} decompose each view into two low-rank matrices with some specific constraints and conduct a conventional clustering approach to group objects. As one of the most significant learning paradigms, canonical correlation analysis (CCA) \cite{CCA} projects two views to a compact collective feature domain where the two views' linear correlation is maximal.

With the development of deep learning, a variety of deep multi-view clustering methods have been proposed recently. Andrew et al. \cite{DCCA} try to search for linearly correlated representation by learning nonlinear transformations of two views with deep canonical correlation analysis (DCCA). As an improvement of DCCA, Wang et al. \cite{DCCAE} add autoencoder-based terms to stimulate the model performance. To resolve the bottleneck of the above two techniques that can only be applied to two views, Benton et al. \cite{DGCCA} further propose to learn a compact representation from data covering more than two views. More recently, Xie et al. \cite{DMJCS} introduce two deep multi-view joint clustering models, in which multiple latent embedding, weighted multi-view learning mechanism and clustering prediction can be learned simultaneously. Xu et al. \cite{DEMVC} adopt collaborative training strategy and alternately share the auxiliary distribution to achieve consistent multi-view clustering assignment. Zhang et al. \cite{AE2-Nets} carefully design a nested autoencoder to incorporate information from heterogeneous sources into a complete representation, which flexibly balances the consistency and complementarity among multiple views. Wen et al. \cite{CDIMC-net} combine view-specific deep feature extractor and graph embedding strategy together to capture robust feature and local structure for each view.

\subsection{Semi-Supervised Clustering}
As is known that semi-supervised learning is a learning paradigm between unsupervised learning and supervised learning that has the ability to jointly use both labeled and unlabeled patterns. It usually appears in machine learning tasks such as regression, classification and clustering. In semi-supervised clustering, pairwise constraints are frequently utilized as a priori knowledge to guide the training procedure, since the pairwise constraints are easy to obtain practically and flexible for scenarios where the number of clusters is inaccessible. In fact, the pairwise constraints can be vividly represented as ``must-link'' (ML) and ``cannot-link'' (CL) used to record the pairwise relationship between two examples in a given dataset. Over the past few years, semi-supervised clustering with pairwise constraints has become an alive area of research. For instance, the literature \cite{semi-KM 1, semi-KM 2} improve classical $K$-means by integrating pairwise constraints. Based on the idea of modifying the similarity matrix, Kamvar et al. \cite{semi-SC} incorporate constraints into spectral clustering (SC) \cite{SC} such that both ML and CL can be well satisfied. Chang et al. \cite{DAC} propose to reestablish the clustering task as a binary pairwise-classification problem, showing excellent clustering results on six image datasets. Shi et al. \cite{ConPaC} utilize pairwise constraints to meet an enhanced performance in face clustering scenario. Wang et al. \cite{SSFPC} conceive soft pairwise constraints to cooperate with fuzzy clustering.

In multi-view learning territory, there are also various pairwise constraints based semi-supervised applications. Tang et al. \cite{CTRL} elaborate a semi-supervised multi-view subspace clustering approach to foster representation learning with the help of a novel regularizer. Nie et al. \cite{MLAN} simultaneously execute multi-view clustering and local structure uncovering in a semi-supervised fashion to learn the local manifold structure of data, achieving a satisfactory clustering performance. Qin et al. \cite{SSSL-M} achieve a desirable shared affinity matrix to realize semi-supervised subspace learning by jointly learning the multiple affinity matrices, the encoding mappings, the latent representation and the block-diagonal structure-induced shared affinity matrix. Bai et al. \cite{SC-MPI} incorporate multi-view constraints to mitigate the influence of inexact constraints from a certain specific view to discover an ideal clustering effectiveness.
Due to space limitations, we refer interested readers to \cite{for readers 1, for readers 2} for a comprehensive understanding.

\begin{figure*}
	\setlength{\abovecaptionskip}{0pt}
	\setlength{\belowcaptionskip}{0pt}
	\renewcommand{\figurename}{Figure}
	\centering
	\includegraphics[width=0.9\textwidth]{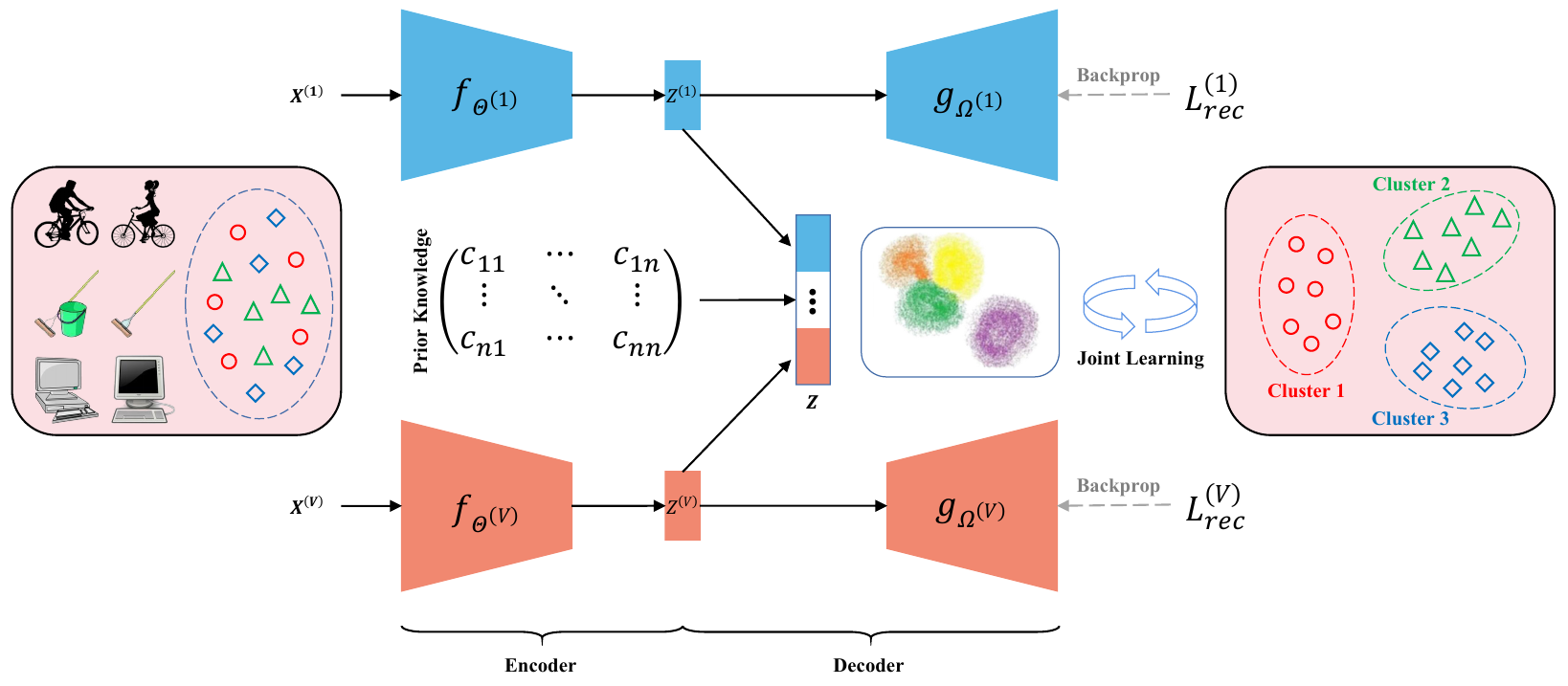}
	\caption{The overall framework of the proposed DMSC approach.}
	\label{structure}
\end{figure*}

\section{Methodology} \label{section:Methodology}
This section elaborates the proposed Deep Multi-view Semi-supervised Clustering (DMSC). Suppose one multi-view dataset $\mathbf{X} = \{\mathbf{X}^{(v)}\}_{v=1}^{V}$ with $V$ views provided, we use $\mathbf{X}^{(v)} = \{\mathbf{x}_{1}^{(v)}, \mathbf{x}_{2}^{(v)}, \ldots, \mathbf{x}_{n}^{(v)}\} \in \mathbb{R}^{D^{(v)} \times n}$ to represent the sample set of the $v$-th view, where $D^{(v)}$ is the feature dimension and $n$ denotes the number of unlabeled instances. Given a little prior knowledge of pairwise constraints, we construct a sparse symmetric matrix $\mathbf{C} = (c_{ik})_{n \times n}$ ($0 \leq i, k \leq n$) with its diagonal elements all zero to describe the connection relationship between pairwise patterns. If pairwise examples share the same label, a ML constraint is built, i.e., $c_{ik} = c_{ki} = 1$ ($i \neq k$), and $c_{ik} = c_{ki} = -1$ ($i \neq k$) otherwise, generating a CL constraint. Provided that the number of clusters $K$ is predefined according to the ground-truth, our goal is to cluster these $n$ multi-view patterns into $K$ groups using prior information $\mathbf{C}$, and we also wish that points with the same label are near to each other, while points from different categories are far away from each other. The overall framework of our DMSC is portrayed in Figure \ref{structure}.

\subsection{Parameters Initialization}
Similar to some previous studies \cite{DMJCS, DEMVC, IDEC, DCN, ASPC}, the proposed model also needs pretraining for a better clustering initialization. In our proposal, we utilize heterogeneous autoencoders as different deep branches to efficiently extract the view-specific feature for every independent view. Specifically, in the $v$-th view, each sample $\mathbf{x}_{i}^{(v)} \in \mathbb{R}^{D^{(v)}}$ is first transformed to a $d^{(v)}$-dimensional feature space by the encoder network $f_{\bm{\Theta}^{(v)}}(\cdot)$:

\begin{equation}
\begin{aligned}
	\mathbf{z}_{i}^{(v)} = f_{\bm{\Theta}^{(v)}}(\mathbf{x}_{i}^{(v)}),
\end{aligned}
\end{equation}
and then is reconstructed by the decoder network $g_{\bm{\Omega}^{(v)}}(\cdot)$ using the corresponding $d^{(v)}$-dimensional latent embedding $\mathbf{z}_{i}^{(v)} \in \mathbb{R}^{d^{(v)}}$:

\begin{equation}
\begin{aligned}
	\hat{\mathbf{x}}_{i}^{(v)} = g_{\bm{\Omega}^{(v)}}(\mathbf{z}_{i}^{(v)}),
\end{aligned}
\end{equation}
where $d^{(v)} \ll D^{(v)}$. Obviously, in an unsupervised mode, it is easy to obtain the initial high-level compact representation for view $v$ by minimizing the following loss function:

\begin{equation}
\begin{aligned}
	L_{rec}^{(v)} = \sum_{i=1}^{n} \|\mathbf{x}_{i}^{(v)} - \hat{\mathbf{x}}_{i}^{(v)}\|_{2}^{2}.
\end{aligned}
\end{equation}
Therefore, the total reconstruction loss of all views can be computed by

\begin{equation}
\begin{aligned}
	L_{rec} = \sum_{v=1}^{V} L_{rec}^{(v)}.
\end{aligned}
\end{equation}
After pretraining of multiple deep branches, a familiar treatment is directly concatenating the embedded features $\mathbf{Z}^{(v)} = \{\mathbf{z}_{1}^{(v)}, \mathbf{z}_{2}^{(v)}, \ldots, \mathbf{z}_{n}^{(v)}\} \in \mathbb{R}^{d^{(v)} \times n}$ as $\mathbf{Z} = \{\mathbf{Z}^{(v)}\}_{v=1}^{V}$ and carrying out $K$-means to achieve initialized cluster centers $\mathbf{M} = \{\mathbf{M}^{(v)}\}_{v=1}^{V}$ with $\mathbf{M}^{(v)} = \{\mathbf{m}_{1}^{(v)}, \mathbf{m}_{2}^{(v)}, \ldots, \mathbf{m}_{K}^{(v)}\} \in \mathbb{R}^{d^{(v)} \times K}$.

\subsection{Networks Finetuning with Pairwise Constraints}
In the finetuning stage, the anterior study \cite{DMJCS} introduces a novel multi-view soft assignment distribution to implement the multi-view fusion, which is defined as

\begin{equation}
\begin{aligned}
	q_{ij} = \frac{\sum_{v} \pi_{j}^{(v)}(1 + \|\mathbf{z}_{i}^{(v)} - \mathbf{m}_{j}^{(v)}\|_{2}^{2} / \alpha)^{-\frac{\alpha + 1}{2}}}{\sum_{j'} \sum_{v'} \pi_{j'}^{(v')}(1 + \|\mathbf{z}_{i}^{(v')} - \mathbf{m}_{j'}^{(v')}\|_{2}^{2} / \alpha)^{-\frac{\alpha + 1}{2}}},
\end{aligned}
\end{equation}
where $\pi_{j}^{(v)}$ indicates the importance weight that measures the importance of the cluster center $\mathbf{m}_{j}^{(v)}$ for consistent clustering. As narrated in \cite{DMJCS}, this multi-view soft assignment distribution (denoted as $\mathbf{Q}$) attains the multi-view fusion via implicitly exerting the multi-view constraint on the view-specific soft assignment (denoted as $\mathbf{Q}^{(v)}$), which is more advantageous than single-view one in \cite{DEC}. Note that there are two constraints for $\pi_{j}^{(v)}$, i.e.,

\begin{equation}
\begin{aligned}
	\pi_{j}^{(v)} \geq 0,
\end{aligned}
\end{equation}
and

\begin{equation}
\begin{aligned}
	\sum_{v=1}^{V} \pi_{j}^{(v)} = 1,
\end{aligned}
\end{equation}
It is not hard to notice that directly optimizing the objective function with respect to $\pi_{j}^{(v)}$ is laborious. Therefore, the constrained weight $\pi_{j}^{(v)}$ can be logically represented in terms of the unconstrained weight $w_{j}^{(v)}$ in a softmax like form as

\begin{equation}
\begin{aligned}
	\pi_{j}^{(v)} = \frac{e^{w_{j}^{(v)}}}{\sum_{v'} e^{w_{j}^{(v')}}}.
\end{aligned}
\end{equation}
In this way, $\pi_{j}^{(v)}$ can definitely meet the above two limitations (6)(7) and $w_{j}^{(v)}$ can be expediently learned by stochastic gradient descent (SGD) as well. For simplicity, the view importance matrix $\mathbf{W}$ is constructed to collect $K \times V$ unconstrained weights $w_{j}^{(v)}$ with $K$ and $V$ being the number of clusters and the number of views respectively. To optimize the multi-view soft assignment distribution $\mathbf{Q}$, the auxiliary target distribution $\mathbf{P}$ is further derived as

\begin{equation}
\begin{aligned}
	p_{ij} = \frac{q_{ij}^{2} / \sum_i q_{ij}}{\sum_{j'} q_{ij'}^{2} / \sum_i q_{ij'}}.
\end{aligned}
\end{equation}
The auxiliary target distribution $\mathbf{P}$ can guide the clustering by enhancing the discrimination of the soft assignment distribution $\mathbf{Q}$. As a result, with the help of $\mathbf{Q}$ and $\mathbf{P}$, the KL divergence based clustering loss is defined as

\begin{equation}
\begin{aligned}
	L_{clu} = KL(\mathbf{P} \ \| \ \mathbf{Q}) = \sum_{i=1}^{n} \sum_{j=1}^{K} p_{ij}\text{log}\frac{p_{ij}}{q_{ij}}.
\end{aligned}
\end{equation}

Owning an excellent learning paradigm, DEC \cite{DEC} like multi-view learning methods \cite{DMJCS, DEMVC} take samples with high confidence as supervisory signals to make them more densely distributed in each cluster, which is the main innovation and contribution. However, they fail to take advantage of user-specific pairwise constraints to boost clustering performance. In order to track this issue, drawing lessons from \cite{SDEC}, we innovatively propose to integrate pairwise constraints into the objective (10) to bring about more robust joint multi-view representation learning and latent clustering. As mentioned earlier, the constraint matrix $\mathbf{C}$ is used for storing ML and CL constraints. When the ML constraint is established, a pair of data points share the same cluster, while satisfying the CL constraint means that the pairwise patterns belong to different clusters. Meanwhile, we also hope that this kind of prior information can help the model better force the two instances to be scattered in their correct and reasonable clusters. To achieve this aim, a $\ell_{2}$-norm based semi-supervised loss employed to measure the connection status between sample $i$ and sample $k$ is defined as follows:

\begin{equation}
\begin{aligned}
	L_{con} = \sum_{i=1}^{n} \sum_{k=1}^{n} c_{ik}\|\mathbf{z}_{i} - \mathbf{z}_{k}\|_{2}^{2},
\end{aligned}
\end{equation}
where $\mathbf{z}_{i} = [\mathbf{z}_{i}^{(1)}; \mathbf{z}_{i}^{(2)}; \ldots; \mathbf{z}_{i}^{(V)}]$ and $\mathbf{z}_{k} = [\mathbf{z}_{k}^{(1)}; \mathbf{z}_{k}^{(2)}; \ldots; \mathbf{z}_{k}^{(V)}]$ are the two concatenated feature points. $c_{ik}$ is a scalar variable that always satisfies the following settings:

\begin{equation}
	c_{ik} = \left\{
\begin{array}{lr}
	1, \quad\quad &(\mathbf{x}_{i}, \mathbf{x}_{k}) \in \text{ML},\\
	-1, \quad\quad &(\mathbf{x}_{i}, \mathbf{x}_{k}) \in \text{CL},\\
	0, \quad\quad &\text{otherwise}.
\end{array}
\right.
\end{equation}
By introducing these valuable weak supervisors $\{c_{ik}\}$ ($1 \leq i, k \leq n$), the model could furnish a strong pulling force over data themselves, so that patterns sharing the same ground-truth label can be as crowded as possible, while those with conflict categories are far away from each other. In reality, benefiting from this, the formed clustering construction would be more rational and prettily, where the elements lying in the cluster are quite agglomerative and the distances between clusters are far-off enough.

Furthermore, to guard against the corruption of common feature space and to protect view-specific feature properties simultaneously, inspired by \cite{IDEC, DCN}, we further propose retaining the view-specific decoders and taking their reconstruction losses into account during network finetuning. Naturally, with the reconstruction part considered, a more robust shared representation can be learned to create a stable training process and a cluster-friendly circumstance. In summary, the objective of our enhanced model, called Deep Multi-view Semi-supervised Clustering (DMSC), can be formulated as

\begin{equation}
\begin{aligned}
	L = L_{rec} + \gamma \cdot L_{clu} + \lambda \cdot L_{con},
\end{aligned}
\end{equation}
where $L_{rec}$ indicates the total reconstruction loss of multiple deep autoencoders as Eq. (4). $L_{clu}$ refers to the KL divergence between multi-view soft assignment distribution $\mathbf{Q}$ and auxiliary target distribution $\mathbf{P}$. $L_{con}$ represents the aforementioned semi-supervised pairwise constraint loss. $\gamma$ and $\lambda$ are two balance factors attached on $L_{clu}$ and $L_{con}$ respectively to trade off the three terms of losses.

As a matter of fact, optimizing Eq. (13) brings two benefits: 1) the costs of violated constraints can be minimized to generate a more reasonable cluster-oriented architecture; 2) both the local structure of specific view feature and the common global attributes of multiple view features can be well preserved so as to perform a better clustering achievement. These two superiorities lead our model to be able to jointly learn a shared high-quality representation and perform a perfect clustering assignment in a semi-supervised manner based on the user-given prior knowledge.

\subsection{Optimization}
In this subsection, we focus on the optimization in the finetuning stage, where mini-batch stochastic gradient decent (SGD) and backpropagation (BP) are resorted to optimize the loss function (13). Specifically, there are four types of variables need to be updated: network parameters $\bm{\Theta}^{(v)}$, $\bm{\Omega}^{(v)}$, cluster center $\mathbf{m}_{j}^{(v)}$, unconstrained importance weight $w_{j}^{(v)}$ and target distribution $\mathbf{P}$. Note that the constrained importance weight $\pi_{j}^{(v)}$ is initialized as $1/V$ and the initial network parameters $\bm{\Theta}^{(v)}$, $\bm{\Omega}^{(v)}$ are gained by pretraining isomeric network branches (i.e., by minimizing Eq. (3) for each view).

\subsubsection{Update $\mathbf{\Theta}^{(v)}$, $\mathbf{\Omega}^{(v)}$, $\mathbf{m}_{j}^{(v)}$, $w_{j}^{(v)}$}
With target distribution $\mathbf{P}$ fixed, the gradients of $L_{clu}$ with respect to feature point $\mathbf{z}_{i}^{(v)}$, cluster center $\mathbf{m}_{j}^{(v)}$, and unconstrained importance weight $w_{j}^{(v)}$ for the $v$-th view are respectively computed as

\begin{equation}
\begin{aligned}
	\frac{\partial L_{clu}}{\partial \mathbf{z}_{i}^{(v)}} = \frac{2}{\alpha} \times \sum_{j=1}^{K} \frac{\partial L_{clu}}{\partial d_{ij}^{(v)}}(\mathbf{z}_{i}^{(v)} - \mathbf{m}_{j}^{(v)}),
\end{aligned}
\end{equation}
\begin{equation}
\begin{aligned}
	\frac{\partial L_{clu}}{\partial \mathbf{m}_{j}^{(v)}} = -\frac{2}{\alpha} \times \sum_{i=1}^{n} \frac{\partial L_{clu}}{\partial d_{ij}^{(v)}}(\mathbf{z}_{i}^{(v)} - \mathbf{m}_{j}^{(v)}),
\end{aligned}
\end{equation}
\begin{equation}
\begin{aligned}
	\frac{\partial L_{clu}}{\partial w_{j}^{(v)}} = \pi_{j}^{(v)}(\frac{\partial L_{clu}}{\partial \pi_{j}^{(v)}} - \sum_{v'=1}^{V} \pi_{j}^{(v')}\frac{\partial L_{clu}}{\partial \pi_{j}^{(v')}}),
\end{aligned}
\end{equation}
where $d_{ij}^{(v)} = \|\mathbf{z}_{i}^{(v)} - \mathbf{m}_{j}^{(v)}\|_{2}^{2} / \alpha$ can be described as the distance between $\mathbf{z}_{i}^{(v)}$ and $\mathbf{m}_{j}^{(v)}$. Let

\begin{equation}
\begin{aligned}
	u = \sum_{j'=1}^{K} \sum_{v'=1}^{V} \pi_{j'}^{(v')}(1 + d_{ij'}^{(v')})^{-\frac{\alpha + 1}{2}},
\end{aligned}
\end{equation}
since $\alpha = 1.0$ set in \cite{DEC}, thus the gradient derivations of $\frac{\partial L_{clu}}{\partial d_{ij}^{(v)}}$ and $\frac{\partial L_{clu}}{\partial \pi_{j}^{(v)}}$ are

\begin{equation}
\begin{aligned}
	\frac{\partial L_{clu}}{\partial d_{ij}^{(v)}} = \frac{\pi_{j}^{(v)}(1 + d_{ij}^{(v)})^{-1}}{q_{ij}u}(p_{ij} - q_{ij})(1 + d_{ij}^{(v)})^{-1},
\end{aligned}
\end{equation}
\begin{equation}
\begin{aligned}
	\frac{\partial L_{clu}}{\partial \pi_{j}^{(v)}} = -\sum_{i=1}^{n} \frac{p_{ij}}{q_{ij}u}(1 - q_{ij})(1 + d_{ij}^{(v)})^{-1}.
\end{aligned}
\end{equation}
Similarly, it is easy to prove that the gradients of $L_{con}$ with respect to $\mathbf{z}_{i} = \{\mathbf{z}_{i}^{(v)}\}_{v=1}^{V}$ can be expressed as follows:

\begin{equation}
\begin{aligned}
	\frac{\partial L_{con}}{\partial \mathbf{z}_{i}} = 2 \times \sum_{k=1}^{n} c_{ik}(\mathbf{z}_{i} - \mathbf{z}_{k}).
\end{aligned}
\end{equation}

It is evidently clear that the gradients $\frac{\partial L_{clu}}{\partial \mathbf{z}_{i}^{(v)}}$ and $\frac{\partial L_{con}}{\partial \mathbf{z}_{i}}$ ($\frac{\partial L_{con}}{\partial \{\mathbf{z}_{i}^{(v)}\}_{v=1}^{V}}$) can be passed down to the corresponding deep network to further compute $\frac{\partial L_{clu}}{\partial \bm{\Theta}^{(v)}}$ and $\frac{\partial L_{con}}{\partial \bm{\Theta}^{(v)}}$ during backpropagation (BP). As a result, given a mini-batch with $m$ samples and learning rate $\eta$, the network parameters $\bm{\Theta}^{(v)}$, $\bm{\Omega}^{(v)}$ are updated by

\begin{equation}
\begin{aligned}
	\bm{\Theta}^{(v)*} = \bm{\Theta}^{(v)} - \frac{\eta}{m}\sum_{i=1}^{m} (\frac{\partial L_{rec}}{\partial \bm{\Theta}^{(v)}} + \gamma \cdot \frac{\partial L_{clu}}{\partial \bm{\Theta}^{(v)}} + \lambda \cdot \frac{\partial L_{con}}{\partial \bm{\Theta}^{(v)}}),
\end{aligned}
\end{equation}
\begin{equation}
\begin{aligned}
	\bm{\Omega}^{(v)*} = \bm{\Omega}^{(v)} - \frac{\eta}{m}\sum_{i=1}^{m} \frac{\partial L_{rec}}{\partial \bm{\Omega}^{(v)}}.
\end{aligned}
\end{equation}
The cluster center $\mathbf{m}_{j}^{(v)}$ and the unconstrained importance weight $w_{j}^{(v)}$ are updated by

\begin{equation}
\begin{aligned}
	\mathbf{m}_{j}^{(v)*} = \mathbf{m}_{j}^{(v)} - \frac{\eta}{m}\sum_{i=1}^{m} \frac{\partial L_{clu}}{\partial \mathbf{m}_{j}^{(v)}},
\end{aligned}
\end{equation}
\begin{equation}
\begin{aligned}
	w_{j}^{(v)*} = w_{j}^{(v)} - \frac{\eta}{m}\sum_{i=1}^{m} \frac{\partial L_{clu}}{\partial w_{j}^{(v)}}.
\end{aligned}
\end{equation}

\subsubsection{Update $\mathbf{P}$}
Although the target distribution $\mathbf{P}$ serves as a ground-truth soft label to facilitate clustering, it also depends on the predicted soft assignment $\mathbf{Q}$. Hence, we should not update $\mathbf{P}$ at each iteration just using a mini-batch of samples to avoid numerical instability. In practice, $\mathbf{P}$ should be updated considering all embedded feature points every $U$ iterations. The update interval $U$ is determined jointly by both sample size $n$ and mini-batch size $m$. After $\mathbf{P}$ is updated, the pseudo label $s_{i}$ for sample $\mathbf{x}_{i} = \{\mathbf{x}_{i}^{(v)}\}_{v=1}^{V}$ is obtained by

\begin{equation}
\begin{aligned}
	s_{i} = \arg \max_{j} \mathbf{q}_{i} \ (j = 1, 2, \ldots, K).
\end{aligned}
\end{equation}

\subsubsection{Stopping Criterion}
If the change in predicted pseudo labels between two consecutive update intervals $U$ is not greater than a threshold $\delta$, we will terminate the training procedure. Formally, the stopping criterion can be written as

\begin{equation}
\begin{aligned}
	1 - \frac{1}{n}\sum_{i}^{n} \sum_{j}^{K} g^{2 \times U}_{ij} g^{1 \times U}_{ij} \leq \delta,
\end{aligned}
\end{equation}
where $g^{2 \times U}_{ij}$ and $g^{1 \times U}_{ij}$ are indicators for whether the $i$-th example is clustered to the $j$-th group at the $(2 \times U)$-th and $(1 \times U)$-th iteration, respectively. We empirically set $\delta = 10^{-4}$ in our subsequent experiments.

\begin{algorithm}
	\SetAlgoLined
	\caption{\textbf{Deep Multi-view Semi-supervised Clustering with Sample Pairwise Constraints}}
	\label{algo}
	\KwIn{Dataset $\mathbf{X}$; Number of clusters $K$; Maximum iterations $T$; Update interval $U$; Stopping threshold $\delta$; Degree of freedom $\alpha$; Proportion of prior knowledge $\beta$; Parameters $\gamma$ and $\lambda$.}
	\KwOut{Clustering assignment $\mathbf{S}$.}
	\BlankLine
	// \textit{\textbf{Initialization}}\\
	Initialize $\mathbf{C}$ by (12), (30);\\
	Initialize $\bm{\Theta}^{(v)}$, $\bm{\Omega}^{(v)}$ by minimizing (3);\\
	Initialize $\mathbf{M}^{(v)} = \{\mathbf{m}_{j}^{(v)}\}_{j=1}^{K}$, $\mathbf{S} = \{s_{i}\}_{i=1}^{n}$ by performing $K$-means on $\mathbf{Z}^{(v)} = \{\mathbf{z}_{i}^{(v)}\}_{i=1}^{n}$;\\
	\BlankLine
	// \textit{\textbf{Finetuning}}\\
	\For {$t \in \{0, 1, \ldots, T-1\}$}
	{
		Select a mini-batch $\mathbf{B}^{(v)} \subset \mathbf{X}^{(v)}$ with $m$ samples and set the learning rate as $\eta$;\\
		\If {$t \% U == 0$}
		{
			Update $\mathbf{P}$ by (5), (9);\\
			Update $\mathbf{S}$ by (25);
		}
		\If {Stopping criterion (26) is met}
		{
			Terminate training.
		}
		Update $\bm{\Theta}^{(v)}$, $\bm{\Omega}^{(v)}$ by (21), (22);\\
		Update $\mathbf{M}^{(v)}$ by (23);\\
		Update $\mathbf{w}^{(v)}$ by (24);\\
	}
\end{algorithm}

The entire optimization process is summarized in Algorithm \ref{algo}. By iteratively updating the above variables, the proposed DMSC can converge to the local optimal solution in theory.

\section{Experiment}\label{section:Experiment}
In this section, we carry out comprehensive experiments to investigate the performance of our DMSC. All experiments are implemented on a standard Linux Server with an Intel(R) Xeon(R) Gold 6226R CPU @ 2.90 GHz, 376 GB RAM, and two NVIDIA A40 GPUs (48 GB caches).

\subsection{Datasets}
Eight popular image datasets are employed in our experiments, including USPS \footnote{http://www.cad.zju.edu.cn/home/dengcai/Data/MLData.html}, COIL20 \footnote{https://www.cs.columbia.edu/CAVE/software/softlib/coil-20.php}, MEDICAL \footnote{https://github.com/apolanco3225/Medical-MNIST-Classification}, FASHION \footnote{https://github.com/zalandoresearch/fashion-mnist} and STL10 \footnote{https://cs.stanford.edu/\~{}acoates/stl10/}, COIL100 \footnote{https://www.cs.columbia.edu/CAVE/software/softlib/coil-100.php}, CALTECH101 \footnote{http://www.vision.caltech.edu/Image\_Datasets/Caltech101/}, CIFAR10 \footnote{http://www.cs.toronto.edu/\~{}kriz/cifar.html}.

\begin{itemize}
\item USPS consists of $9298$ grayscale handwritten digit images with a size of $16 \times 16$ pixels from $10$ categories.
\item COIL20 includes $1440$ $128 \times 128$ gray object images from $20$ categories, which are shotted from different angles. The resized version of $32 \times 32$ is adopted in our experiments.
\item MEDICAL is a simple medical dataset in $64 \times 64$ dimension. There are $58954$ medical images belonging to $6$ classes, i.e., abdomen CT, breast MRI, chest X-ray, chest CT, hand X-ray, head CT.
\item FASHION is a collection of $70000$ fashion product images from $10$ classes, with $28 \times 28$ image size and one image channel.
\item STL10 embraces $13000$ color images with the size of $96 \times 96$ pixels from $10$ object categories.
\item COIL100 incorporates $7200$ image samples of $100$ object categories, with $128 \times 128$ image size and three image channels.
\item CALTECH101 owns $8677$ irregular object images from $101$ classes, which is widely utilized in the field of multi-view learning.
\item CIFAR10 comprises $60000$ RGB images of $10$ object classes, whose image size is standardized as $32 \times 32$.
\end{itemize}

\begin{figure*}
	\setlength{\abovecaptionskip}{0pt}
	\setlength{\belowcaptionskip}{0pt}
	\renewcommand{\figurename}{Figure}
	\centering
	\begin{minipage}{0.2\linewidth}
		\centerline{\includegraphics[width=\textwidth]{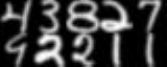}}
		\vspace{0pt}
		\centerline{(a) USPS}
		\vspace{5pt}
		\centerline{\includegraphics[width=\textwidth]{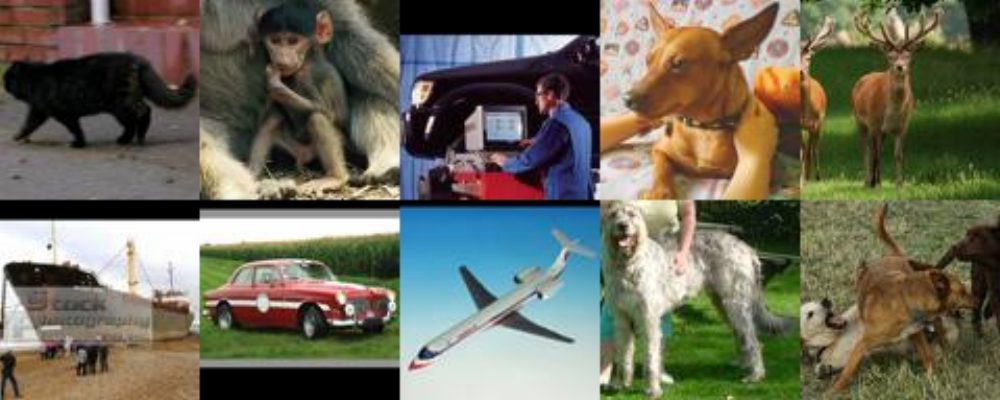}}
		\vspace{0pt}
		\centerline{(e) STL10}
		\vspace{5pt}
	\end{minipage}
	\hspace{3pt}
	\begin{minipage}{0.2\linewidth}
		\centerline{\includegraphics[width=\textwidth]{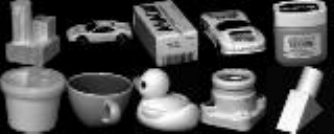}}
		\vspace{0pt}
		\centerline{(b) COIL20}
		\vspace{5pt}
		\centerline{\includegraphics[width=\textwidth]{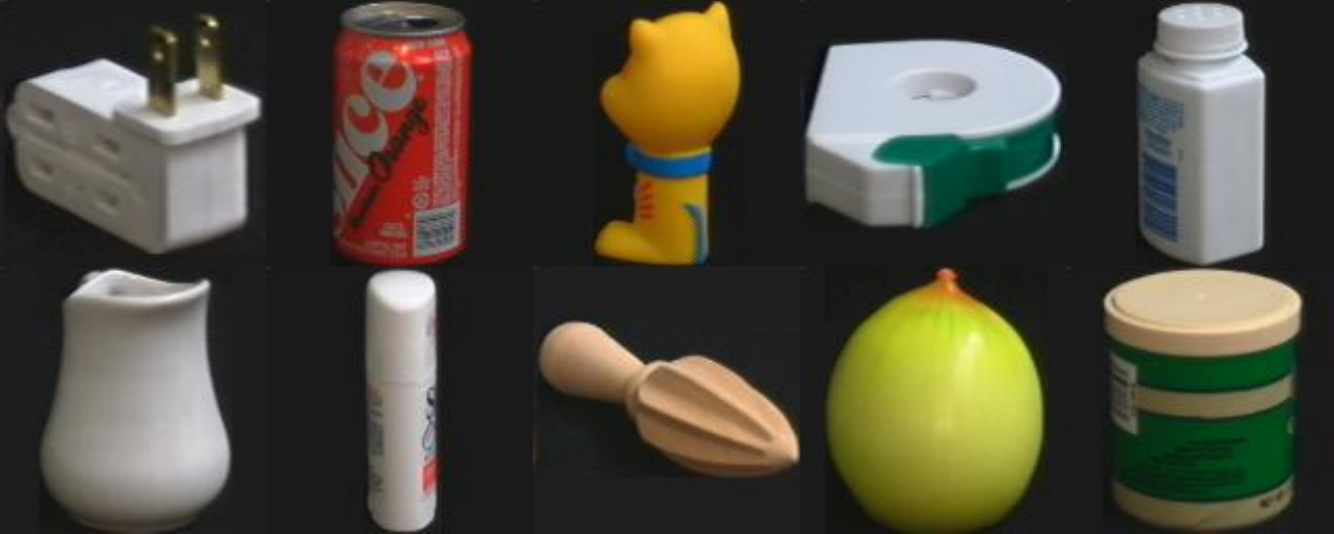}}
		\vspace{0pt}
		\centerline{(f) COIL100}
		\vspace{5pt}
	\end{minipage}
	\hspace{3pt}
	\begin{minipage}{0.2\linewidth}
		\centerline{\includegraphics[width=\textwidth]{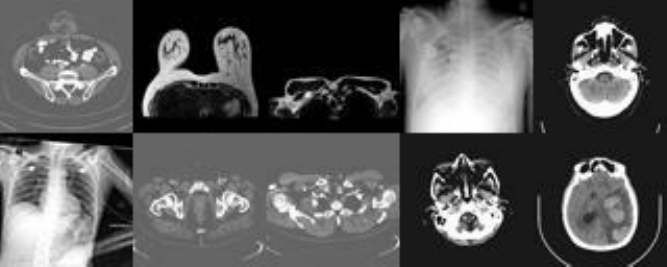}}
		\vspace{0pt}
		\centerline{(c) MEDICAL}
		\vspace{5pt}
		\centerline{\includegraphics[width=\textwidth]{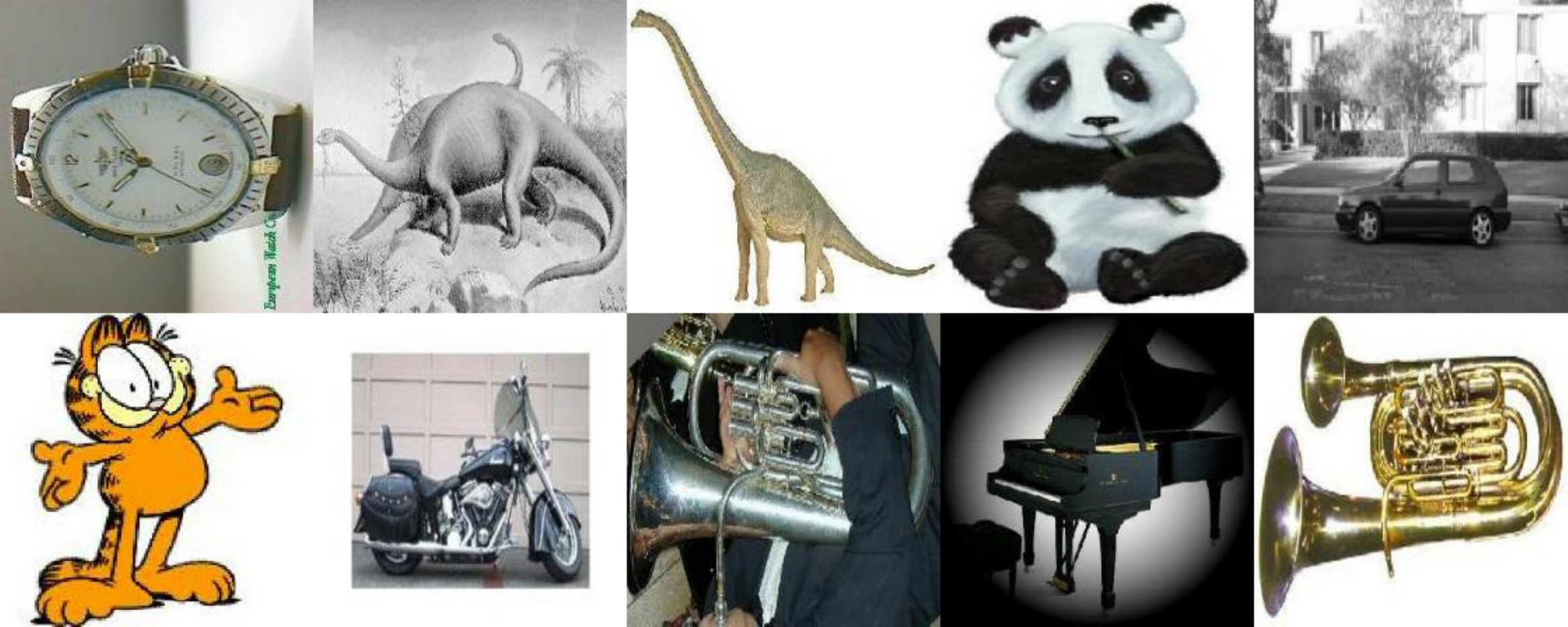}}
		\vspace{0pt}
		\centerline{(g) CALTECH101}
		\vspace{5pt}
	\end{minipage}
	\hspace{3pt}
	\begin{minipage}{0.2\linewidth}
		\centerline{\includegraphics[width=\textwidth]{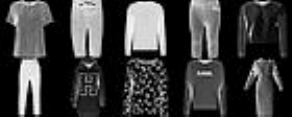}}
		\vspace{0pt}
		\centerline{(d) FASHION}
		\vspace{5pt}
		\centerline{\includegraphics[width=\textwidth]{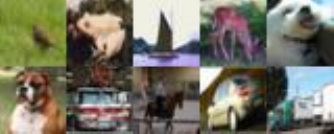}}
		\vspace{0pt}
		\centerline{(h) CIFAR10}
		\vspace{5pt}
	\end{minipage}
	\caption{The examples of datasets.}
	\label{examples of datasets}
\end{figure*}

The properties and examples are summarized in Table \ref{properties of datasets} and Figure \ref{examples of datasets}. Since these datasets have been split into training set and testing set, both subsets are jointly utilized for clustering analysis. Besides, in our experiments, the aforementioned datasets are rescaled to $[-1,1]$ for each entity before being infused to model training.

\begin{table}
	\renewcommand{\arraystretch}{1.25}
	\caption{The properties of datasets.}
	\label{properties of datasets}
	\centering
	\scalebox{0.8}{
		\begin{threeparttable}
			\begin{tabular}{l|ccccc}
				\hline
				Dataset				&Instance	&Category	&Size				&Channel\\
				\hline
				USPS				&$9298$	&$10$		&$16 \times 16$	&$1$\\
				COIL20			&$1440$	&$20$		&$32 \times 32$	&$1$\\
				MEDICAL			&$58954$	&$6$		&$64 \times 64$	&$1$\\
				FASHION			&$70000$	&$10$		&$28 \times 28$	&$1$\\
				\hline
				STL10				&$13000$	&$10$		&$96 \times 96$	&$3$\\
				COIL100			&$7200$	&$100$	&$128 \times 128$	&$3$\\
				CALTECH101		&$8677$	&$101$	&$-$				&$3$\\
				CIFAR10			&$60000$	&$10$		&$32 \times 32$	&$3$\\
				\hline
			\end{tabular}
			\footnotesize
	\end{threeparttable}}
\end{table}

\subsection{Evaluation Metrics}
We adopt three standard metrics, i.e., clustering accuracy (ACC) \cite{ACC}, normalized mutual information (NMI) \cite{NMI} and adjusted rand index (ARI) \cite{ARI}, to evaluate the performance of different clustering methods. Their definitions can be formulated as follows:

\begin{equation}
\begin{aligned}
	\text{ACC} = \max_{p} \ \frac{\sum_{i=1}^{n} \mathbf{1}\{y_i = p(s_i)\}}{n},
\end{aligned}
\end{equation}
where $n$ is the sample size. $y_i$ and $s_i$ denote the ground-truth label and the clustering assignment generated by the model for the $i$-th pattern, respectively. $p(\cdot)$ is the permutation function, which embraces all possible one-to-one projections from clusters to labels. The best projection can be efficiently computed by the Hungarian \cite{Hungarian algorithm} algorithm.

\begin{equation}
\begin{aligned}
	\text{NMI} = \frac{I(y;s)}{\max\{H(y),H(s)\}},
\end{aligned}
\end{equation}
where $I(y;s)$ and $H$ represent the mutual information between $y$ and $s$ and the entropic cost, respectively.

\begin{equation}
\begin{aligned}
	\text{ARI} = \frac{\text{RI} - E(\text{RI})}{\max(\text{RI}) - E(\text{RI})},
\end{aligned}
\end{equation}
where $E(\text{RI})$ is the expectation of the rand index (RI) \cite{RI}.

Note that ACC and NMI range within $[0,1]$, while the range of ARI is $[-1,1]$, and a higher score indicates a better clustering performance. Generally, the aforesaid metrics are extensively considered in various clustering literature \cite{CTRL, JSTC, T-MEK-SPL, DCSPC, DMNEC}. Each one offers pros and cons, but using them together is sufficient to test the effectiveness of the clustering algorithms.

\subsection{Compared Methods}
Several clustering methods are chosen to comprehensively compare with the proposed DMSC, which can be roughly grouped as: 1) \emph{single-view methods}, including autoencoder (AE), deep embedded clustering (DEC) \cite{DEC}, improved deep embedded clustering (IDEC) \cite{IDEC}, deep clustering network (DCN) \cite{DCN}, adaptive self-paced clustering (ASPC) \cite{ASPC}, semi-supervised deep embedded clustering (SDEC) \cite{SDEC}; 2) \emph{multi-view methods}, containing robust multi-view $K$-means clustering (RMKMC) \cite{RMKMC}, multi-view self-paced clustering (MSPL) \cite{MSPL}, deep canonical correlation analysis (DCCA) \cite{DCCA}, deep canonically correlated autoencoders (DCCAE) \cite{DCCAE}, deep generalized canonical correlation analysis (DGCCA) \cite{DGCCA}, deep multi-view joint clustering with soft assignment distribution (DMJCS) \cite{DMJCS}, deep embedded multi-view clustering (DEMVC) \cite{DEMVC}.

\begin{table*}
	\renewcommand{\arraystretch}{1.25}
	\caption{The experimental configurations.}
	\label{exp config}
	\centering
	\scalebox{0.8}{
		\begin{threeparttable}
			\begin{tabular}{l|l|l|l}
				\hline
				Dataset		&Branch				&Encoder																						&Input\\
				\hline
				1-channel	&View 1 (SAE)		&$\text{Fc}_{500}-\text{Fc}_{500}-\text{Fc}_{2000}-\text{Fc}_{10}$						&Raw image vectors\\
				{}			&View 2 (CAE)		&$\text{Conv}_{32}^{5}-\text{Conv}_{64}^{5}-\text{Conv}_{128}^{3}-\text{Fc}_{10}$	&Raw image pixels\\
				\hline
				3-channel	&View 1 (SAE)		&$\text{Fc}_{500}-\text{Fc}_{500}-\text{Fc}_{2000}-\text{Fc}_{10}$						&DenseNet121 feature\\
				{}			&View 2 (SAE)		&$\text{Fc}_{500}-\text{Fc}_{256}-\text{Fc}_{50}$											&InceptionV3 feature\\
				\hline
			\end{tabular}
			\footnotesize
	\end{threeparttable}}
\end{table*}

\begin{table*}
	\renewcommand{\arraystretch}{1.25}
	\caption{The experimental comparison on grayscale image datasets.}
	\label{experimental comparison on 1c}
	\centering
	\scalebox{0.7}{
		\begin{threeparttable}
			\begin{tabular}{l|l|ccc|ccc|ccc|ccc}
				\hline
				\multirow{2}{*}{Type}&\multirow{2}{*}{Method}&\multicolumn{3}{c|}{USPS}&\multicolumn{3}{c|}{COIL20}&\multicolumn{3}{c|}{MEDICAL}&\multicolumn{3}{c}{FASHION}\\
				{}		&{}							&ACC&NMI&ARI			&ACC&NMI&ARI			&ACC&NMI&ARI			&ACC&NMI&ARI\\
				\hline
				SvC	&AE-View1					&0.7198&0.7036&0.6017	&0.5643&0.7206&0.5101	&0.6513&0.7157&0.5603	&0.5862&0.5899&0.4522\\
				{}		&AE-View2					&0.7388&0.7309&0.6506	&0.6729&0.7676&0.5986	&0.7208&0.8180&0.7015	&0.6250&0.6476&0.4925\\
				{}		&AE-View1,2					&0.7425&0.7413&0.6613	&0.6764&0.7751&0.6037	&0.7227&0.8206&0.7030	&0.6268&0.6496&0.4961\\
				{}		&DEC \cite{DEC}				&0.7532&0.7544&0.6852	&0.5756&0.7650&0.5555	&0.6633&0.7300&0.5842	&0.5923&0.6076&0.4632\\
				{}		&IDEC \cite{IDEC}			&0.7680&0.7794&0.7080	&0.5990&0.7702&0.5745	&0.6836&0.7753&0.6358	&0.5977&0.6348&0.4740\\
				{}		&DCN \cite{DCN}				&0.7367&0.7353&0.6354	&0.5982&0.7463&0.5430	&0.6670&0.7350&0.5723	&0.5947&0.6329&0.4678\\
				{}		&ASPC \cite{ASPC}			&0.7578&0.7673&0.6753	&0.5935&0.7644&0.5535	&0.6960&0.7692&0.6155	&0.6036&0.6385&0.4806\\
				{}		&SDEC \cite{SDEC}			&0.7630&0.7705&0.6995	&0.5915&0.7717&0.5650	&0.6748&0.7466&0.6055	&0.6028&0.6243&0.4754\\
				\hline
				MvC	&RMKMC \cite{RMKMC}		&0.7441&0.7278&0.6667	&0.5799&0.7487&0.5275	&$-$&$-$&$-$				&0.5912&0.6169&0.4636\\
				{}		&MSPL \cite{MSPL}			&0.7414&0.7174&0.6370	&0.5992&0.7623&0.5608	&$-$&$-$&$-$				&0.5607&0.6068&0.4457\\
				{}		&DCCA \cite{DCCA}			&0.4042&0.3895&0.2480	&0.5512&0.7013&0.4600	&$-$&$-$&$-$				&0.4105&0.4028&0.2342\\
				{}		&DCCAE \cite{DCCAE}		&0.3793&0.3895&0.2135	&0.5551&0.7058&0.4667	&$-$&$-$&$-$				&0.4109&0.3836&0.2303\\
				{}		&DGCCA \cite{DGCCA}		&0.5473&0.5079&0.4011	&0.5337&0.6762&0.4370	&$-$&$-$&$-$				&0.4765&0.4827&0.3105\\
				{}		&DMJCS \cite{DMJCS}		&0.7727&0.7941&0.7207	&0.6986&0.8001&0.6384	&0.7341&0.7837&0.6737	&0.6370&0.6628&0.5143\\
				{}		&DEMVC \cite{DEMVC}		&0.7803&0.8051&0.7245	&0.7033&0.8049&0.6453	&0.7387&0.8199&0.7006	&0.6357&0.6605&0.5006\\
				{}		&DMSC (ours)					&\textbf{0.7866}&\textbf{0.8163}&\textbf{0.7380}	&\textbf{0.7126}&\textbf{0.8180}&\textbf{0.6644}	&\textbf{0.7451}&\textbf{0.8287}&\textbf{0.7116}	&\textbf{0.6401}&\textbf{0.6686}&\textbf{0.5183}\\
				\hline
			\end{tabular}
			\footnotesize{Note that both SDEC and DMSC are with the semi-supervised learning paradigm.}
	\end{threeparttable}}
\end{table*}

\subsection{Experimental Setups}
In this subsection, we will introduce the experimental setups in detail, including pretraining setup, prior knowledge utilization and finetuning setup.

\subsubsection{Pretraining Setup}
For one-channel image datasets, we use a stacked autoencoder (SAE) and a convolutional autoencoder (CAE) as two different deep network branches to extract low-dimensional multi-view features. Specifically, the raw image vectors and pixels are fed into SAE and CAE respectively. For three-channel image datasets, two SAEs with different structures and data sources are considered as two multiple branches, whose inputs are the pretrained feature extracted by using DenseNet121 \footnote{https://github.com/fchollet/deep-learning-models/releases/tag/v0.8} and InceptionV3 \footnote{https://github.com/fchollet/deep-learning-models/releases/tag/v0.5} on ILSVRC2012 (ImageNet Large Scale Visual Recognition Competition in 2012), with $1024$ and $2048$ dimensions respectively. During pretraining, the Adam \cite{Adam} optimizer with initial learning rate $0.001$ is utilized to train multi-view branches in an end-to-end fashion for $400$ epochs. The batch size is set as $256$. Moveover, all internal layers of each branch are activated by the ReLU \cite{ReLU} nonlinearity function, and the Xariver \cite{Xariver} method is employed as the layer kernel initializer.

\subsubsection{Prior Knowledge Utilization}
The pairwise constraint matrix $\mathbf{C} = (c_{ik})_{n \times n}$ ($1 \leq i, k \leq n$) is randomly constructed on the basis of the ground-truth labels for each dataset. Thus we indiscriminately pick pairs of data samples from the datasets and put forward a hypothesis: if pairwise patterns share the identical label, a connected constraint is generated, otherwise establishing a disconnected constraint, which is expressed as

\begin{equation}
\left\{
\begin{array}{lr}
	c_{ik} = c_{ki} = 1, \quad\quad &(y_i = y_k, i \neq k),\\
	c_{ik} = c_{ki} = -1, \quad\quad &(y_i \neq y_k, i \neq k),\\
	c_{ik} = c_{ki} = 0, \quad\quad &(i = k).\\
\end{array}
\right.
\end{equation}

Note that the symmetric sparse matrix $\mathbf{C}$ provides us with $n^{2}$ sample constraints in total. Due to its symmetry and sparsity, the number of sample constraints should only be adjusted up to $n^{2} / 2$ at most. Based on such recognition, the scalefactor $\beta$ is prophetically set as $1.0$ in our experiments, which supplies $1.0 \times n$ pairwise constraints (or specifically $2 \times 1.0 \times n$ sample constraints) for the learning model. The sensitivity of $\beta$ will be analyzed and discussed later in Section \ref{subsection:Parameter Analysis}.

\subsubsection{Finetuning Setup}
Different from some preceding studies \cite{DEC, DMJCS, ASPC, SDEC, DBC} that only keep the encoding block retained in their model finetuning stage, we conversely preserve the end-to-end structure (i.e., hold back both encoder and decoder simultaneously) of each branch to protect feature properties for the multi-view data. The entire clustering network is trained for $20000$ epochs by equipping the Adam \cite{Adam} optimizer with default learning rate $\eta = 0.001$. The batch size is fixed to $256$. The importance coefficients for clustering loss and constraint loss are set as $\gamma = 10^{-1}$ and $\lambda = 10^{-6}$, respectively. The threshold in stopping criterion is $\delta = 0.01\%$. The degree of freedom for Student’s t-distribution is assigned as $\alpha = 1.0$. The number of clusters $K$ is hypothetically given as a priori knowledge according to the ground-truth, i.e., $K$ equals to the ground-truth cluster numbers.

The above experimental configurations are summarized in Table \ref{exp config}. Besides, for single-view methods, we take the raw images and the concatenated ImageNet features as the network input when performing on gray and color image datasets respectively. For DCCA \cite{DCCA}, DCCAE \cite{DCCAE}, DGCCA \cite{DGCCA}, we concatenate the multiple latent features gained from their model training and directly perform $K$-means. With regard to RMKMC \cite{RMKMC}, MSPL \cite{MSPL}, the low-dimensional embeddings learned by our pretrained multi-view branches are considered as their multiple inputs. As for DEMVC \cite{DEMVC} and DMJCS \cite{DMJCS}, we set the model configuration to be the same as the corresponding recommended setting. Note that for reasonable estimation, we perform $10$ random restarts for all experiments and report the average results to compare with the others based on Python \emph{3.7} and TensorFlow \emph{2.6.0}.

\begin{table*}
	\renewcommand{\arraystretch}{1.25}
	\caption{The experimental comparison on RGB image datasets.}
	\label{experimental comparison on 3c}
	\centering
	\scalebox{0.7}{
		\begin{threeparttable}
			\begin{tabular}{l|l|ccc|ccc|ccc|ccc}
				\hline
				\multirow{2}{*}{Type}&\multirow{2}{*}{Method}&\multicolumn{3}{c|}{STL10}&\multicolumn{3}{c|}{COIL100}&\multicolumn{3}{c|}{CALTECH101}&\multicolumn{3}{c}{CIFAR10}\\
				{}		&{}							&ACC&NMI&ARI			&ACC&NMI&ARI			&ACC&NMI&ARI			&ACC&NMI&ARI\\
				\hline
				SvC	&AE-View1					&0.7521&0.7218&0.6367	&0.7546&0.9278&0.7473	&0.5088&0.7555&0.4259	&0.5300&0.4384&0.3335\\
				{}		&AE-View2					&0.8716&0.8401&0.8023	&0.7079&0.9152&0.7018	&0.5877&0.8019&0.4636	&0.6586&0.5697&0.4696\\
				{}		&AE-View1,2					&0.9098&0.8706&0.8478	&0.7676&0.9393&0.7706	&0.6096&0.8278&0.4924	&0.6658&0.5883&0.4965\\
				{}		&DEC \cite{DEC}				&0.9574&0.9106&0.9091	&0.7794&0.9459&0.7779	&0.6282&0.8364&0.5261	&0.6744&0.5930&0.5137\\
				{}		&IDEC \cite{IDEC}			&0.9605&0.9150&0.9155	&0.7921&0.9481&0.7955	&0.6373&0.8393&0.5398	&0.6866&0.6072&0.5298\\
				{}		&DCN \cite{DCN}				&0.9318&0.8965&0.8781	&0.7771&0.9399&0.7726	&0.6626&0.8418&0.6022	&0.6828&0.6326&0.5308\\
				{}		&ASPC \cite{ASPC}			&0.9381&0.9061&0.8908	&0.7854&0.9497&0.7869	&0.6729&0.8495&0.6087	&0.6692&0.6162&0.5153\\
				{}		&SDEC \cite{SDEC}			&0.9585&0.9120&0.9115	&0.7942&0.9523&0.8009	&0.6433&0.8450&0.5471	&0.6953&0.6141&0.5353\\
				\hline
				MvC	&RMKMC \cite{RMKMC}		&0.8344&0.8273&0.7635	&$-$&$-$&$-$				&$-$&$-$&$-$				&0.5714&0.4688&0.3679\\
				{}		&MSPL \cite{MSPL}			&0.7414&0.7174&0.6370	&$-$&$-$&$-$				&$-$&$-$&$-$				&0.7156&0.5948&0.5174\\
				{}		&DCCA \cite{DCCA}			&0.8411&0.7477&0.6917	&$-$&$-$&$-$				&$-$&$-$&$-$				&0.4242&0.3385&0.2181\\
				{}		&DCCAE \cite{DCCAE}		&0.8235&0.7273&0.6632	&$-$&$-$&$-$				&$-$&$-$&$-$				&0.3960&0.3226&0.2034\\
				{}		&DGCCA \cite{DGCCA}		&0.8960&0.8218&0.7970	&$-$&$-$&$-$				&$-$&$-$&$-$				&0.4703&0.3577&0.2634\\
				{}		&DMJCS \cite{DMJCS}		&0.9374&0.9063&0.8989	&0.7841&0.9532&0.7991	&0.6998&0.8578&0.7054	&0.7184&0.6188&0.5527\\
				{}		&DEMVC \cite{DEMVC}		&0.9582&0.9121&0.9132	&0.7563&0.9382&0.7626	&0.6719&0.8419&0.6991	&0.6998&0.6351&0.5457\\
				{}		&DMSC (ours)					&\textbf{0.9679}&\textbf{0.9268}&\textbf{0.9305}	&\textbf{0.8077}&\textbf{0.9569}&\textbf{0.8159}	&\textbf{0.7161}&\textbf{0.8593}&\textbf{0.7230}	&\textbf{0.7337}&\textbf{0.6442}&\textbf{0.5712}\\
				\hline
			\end{tabular}
			\footnotesize{Note that both SDEC and DMSC are with the semi-supervised learning paradigm.}
	\end{threeparttable}}
\end{table*}

\subsection{Experimental Comparison}\label{subsection:Experimental Comparison}
Table \ref{experimental comparison on 1c} and Table \ref{experimental comparison on 3c} list the clustering results of the compared baseline methods, where the mark ``$-$'' indicates that the experimental results or codes are unavailable from the corresponding paper, and the boldface refers to the best clustering result. As is illustrated, our DMSC achieves the highest scores in terms of all metrics on all datasets among \emph{Type-MvC}, demonstrating its superiority compared to the state-of-the-art deep multi-view clustering algorithms. In particular, the advantages of DMSC over DMJCS \cite{DMJCS} verifiy that: 1) the feature space protection (FSP) mechanism can help preserve the properties of both view-specific embedding and view-shared representation; 2) the user-offered semi-supervised signals are conducive to forming a more perfect clustering structure.

Moreover, we also compare the proposed DMSC with some advanced single-view methods. The quantitative results are exhibited in \emph{Type-SvC}, where we can notice that the single-view rivals are unable to mine useful complementary information since they can only process one single view, thus leading to a poor performance. In contrast, our DMSC can flexibly handle multi-view information, such that the view-specific feature and the inherent complementary information concealed in different views can be simultaneously learned as a robust global representation to obtain a satisfactory clustering result. Additionally, we also found that, as one of the joint learning based clustering algorithms, the proposed DMSC achieves better performance than the corresponding representation-based approaches (i.e., AE-View1, AE-View2, AE-View1,2) in all cases for all metrics, which clearly demonstrates that combining feature learning with pattern partitioning can provide a more appropriate representation for clustering analysis, implying the progressiveness of the joint optimization criterion.

\begin{table*}
	\renewcommand{\arraystretch}{1.25}
	\caption{The performance of DMSC with different configurations on grayscale image datasets.}
	\label{ablation study 1c}
	\centering
	\scalebox{0.7}{
		\begin{threeparttable}
		\begin{tabular}{l|cc|ccc|ccc|ccc|ccc}
		\hline
		\multirow{2}{*}{Method}&\multirow{2}{*}{SEMI}&\multirow{2}{*}{FSP}&\multicolumn{3}{c|}{USPS}&\multicolumn{3}{c|}{COIL20}&\multicolumn{3}{c|}{MEDICAL}&\multicolumn{3}{c}{FASHION}\\
		{}				&{}			&{}			&ACC&NMI&ARI			&ACC&NMI&ARI			&ACC&NMI&ARI			&ACC&NMI&ARI\\
		\hline
		benchmark		&$\times$		&$\times$		&0.7727&0.7941&0.7207	&0.6986&0.8001&0.6384	&0.7341&0.7837&0.6737	&0.6370&0.6628&0.5143\\
		$-$				&$\checkmark$	&$\times$		&0.7825&0.8087&0.7326	&0.7049&0.8054&0.6458	&0.7363&0.7921&0.6790	&0.6386&0.6647&0.5165\\
		$-$				&$\times$		&$\checkmark$	&0.7780&0.8142&0.7353	&0.7052&0.8176&0.6574	&0.7358&0.8201&0.7033	&0.6349&0.6663&0.5172\\
		DMSC (ours)	&$\checkmark$	&$\checkmark$	&\textbf{0.7866}&\textbf{0.8163}&\textbf{0.7380}	&\textbf{0.7126}&\textbf{0.8180}&\textbf{0.6644}	&\textbf{0.7451}&\textbf{0.8287}&\textbf{0.7116}	&\textbf{0.6401}&\textbf{0.6686}&\textbf{0.5183}\\
		\hline
		\end{tabular}
		\footnotesize
		\end{threeparttable}}
\end{table*}

\begin{table*}
	\renewcommand{\arraystretch}{1.25}
	\caption{The performance of DMSC with different configurations on RGB image datasets.}
	\label{ablation study 3c}
	\centering
	\scalebox{0.7}{
		\begin{threeparttable}
		\begin{tabular}{l|cc|ccc|ccc|ccc|ccc}
		\hline
		\multirow{2}{*}{Method}&\multirow{2}{*}{SEMI}&\multirow{2}{*}{FSP}&\multicolumn{3}{c|}{STL10}&\multicolumn{3}{c|}{COIL100}&\multicolumn{3}{c|}{CALTECH101}&\multicolumn{3}{c}{CIFAR10}\\
		{}				&{}			&{}			&ACC&NMI&ARI			&ACC&NMI&ARI			&ACC&NMI&ARI			&ACC&NMI&ARI\\
		\hline
		benchmark		&$\times$		&$\times$		&0.9374&0.9063&0.8989	&0.7841&0.9532&0.7991	&0.6998&0.8578&0.7054	&0.7184&0.6188&0.5527\\
		$-$				&$\checkmark$	&$\times$		&0.9579&0.9115&0.9102	&0.7962&0.9556&0.8097	&0.7029&0.8588&0.7136	&0.7288&0.6276&0.5587\\
		$-$				&$\times$		&$\checkmark$	&0.9368&0.9123&0.8996	&0.7892&0.9510&0.8010	&0.7054&0.8562&0.7070	&0.7243&0.6263&0.5582\\
		DMSC (ours)	&$\checkmark$	&$\checkmark$	&\textbf{0.9679}&\textbf{0.9268}&\textbf{0.9305}	&\textbf{0.8077}&\textbf{0.9569}&\textbf{0.8159}	&\textbf{0.7161}&\textbf{0.8593}&\textbf{0.7230}	&\textbf{0.7337}&\textbf{0.6442}&\textbf{0.5712}\\
		\hline
		\end{tabular}
		\footnotesize
		\end{threeparttable}}
\end{table*}

\begin{figure}
	\setlength{\abovecaptionskip}{0pt}
	\setlength{\belowcaptionskip}{0pt}
	\renewcommand{\figurename}{Figure}
	\centering
 	\begin{minipage}{0.5\linewidth}
	 	\centerline{\includegraphics[width=\textwidth]{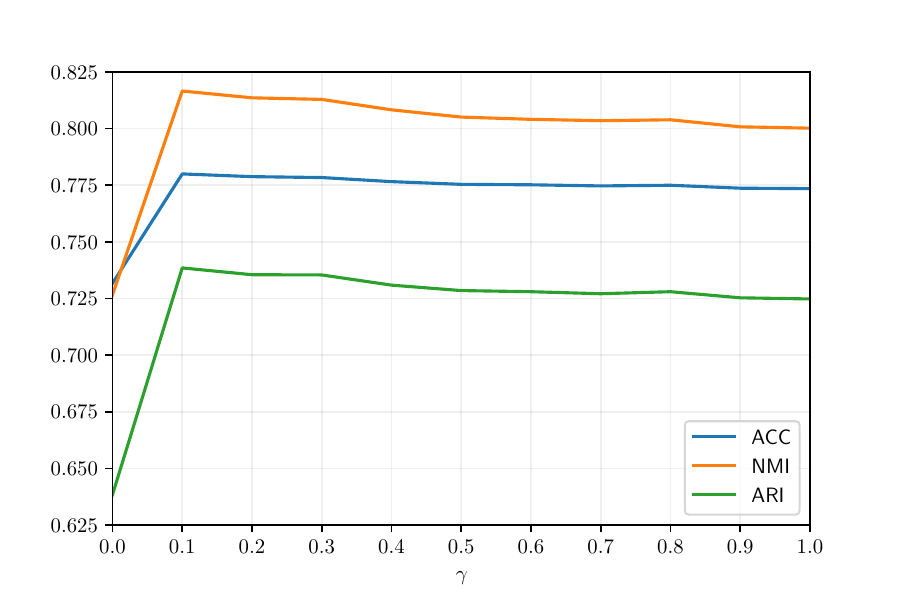}}
		\centerline{(a) USPS}
		\vspace{5pt}
 	\end{minipage}
	\hspace{-15pt}
	\begin{minipage}{0.5\linewidth}
		\centerline{\includegraphics[width=\textwidth]{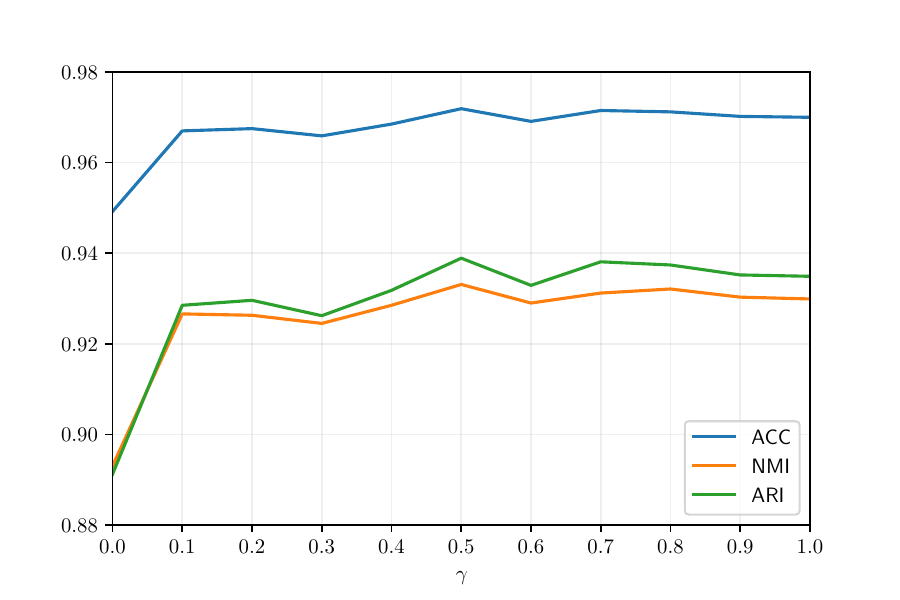}}
		\centerline{(b) STL10}
		\vspace{5pt}
 	\end{minipage}
	\caption{Clustering performance v.s. parameter $\gamma$.}
	\label{parameter analysis of gamma}
\end{figure}

\begin{figure}
	\setlength{\abovecaptionskip}{0pt}
	\setlength{\belowcaptionskip}{0pt}
	\renewcommand{\figurename}{Figure}
	\centering
 	\begin{minipage}{0.5\linewidth}
	 	\centerline{\includegraphics[width=\textwidth]{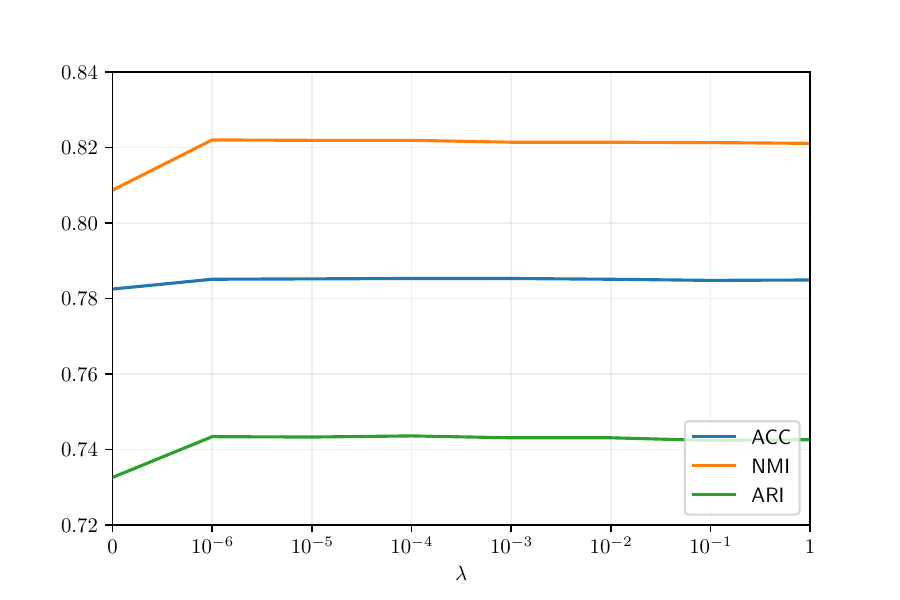}}
		\centerline{(a) USPS}
		\vspace{5pt}
 	\end{minipage}
	\hspace{-15pt}
	\begin{minipage}{0.5\linewidth}
		\centerline{\includegraphics[width=\textwidth]{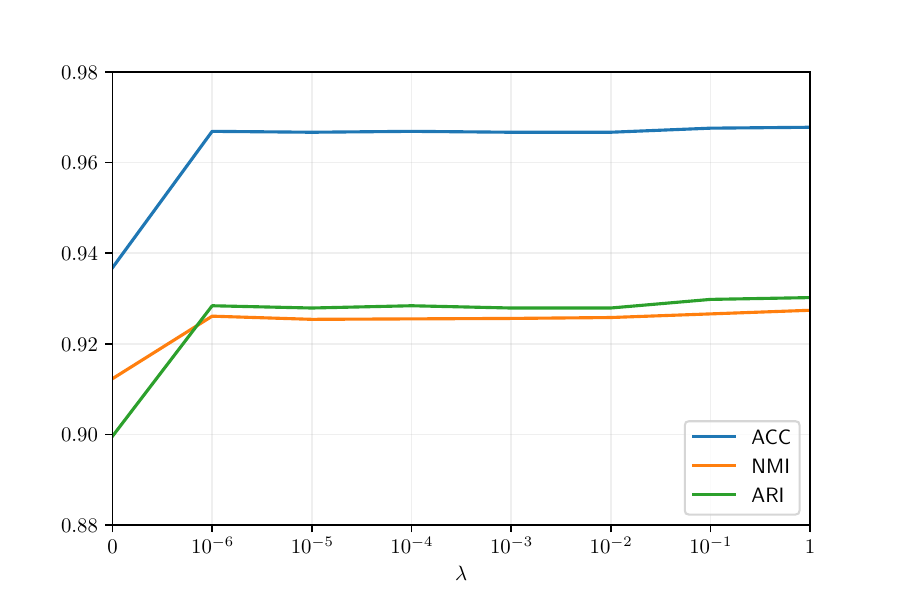}}
		\centerline{(b) STL10}
		\vspace{5pt}
 	\end{minipage}
	\caption{Clustering performance v.s. parameter $\lambda$.}
	\label{parameter analysis of lambda}
\end{figure}

\begin{figure}
	\setlength{\abovecaptionskip}{0pt}
	\setlength{\belowcaptionskip}{0pt}
	\renewcommand{\figurename}{Figure}
	\centering
	\begin{minipage}{0.5\linewidth}
		\centerline{\includegraphics[width=\textwidth]{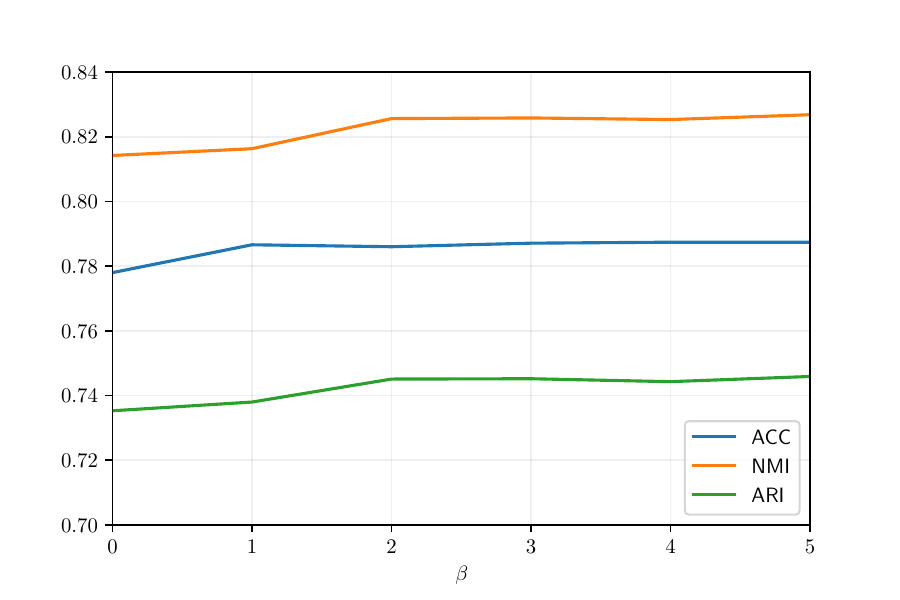}}
		\centerline{(a) USPS}
		\vspace{5pt}
 	\end{minipage}
	\hspace{-15pt}
	\begin{minipage}{0.5\linewidth}
		\centerline{\includegraphics[width=\textwidth]{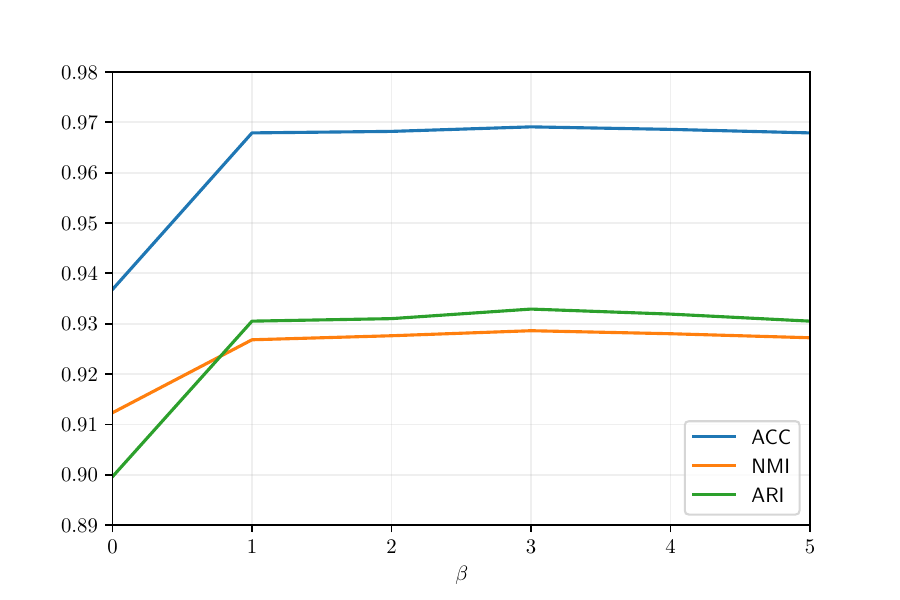}}
		\centerline{(b) STL10}
		\vspace{5pt}
 	\end{minipage}
	\caption{Clustering performance v.s. parameter $\beta$.}
	\label{parameter analysis of beta}
\end{figure}

\begin{figure}
	\setlength{\abovecaptionskip}{0pt}
	\setlength{\belowcaptionskip}{0pt}
	\renewcommand{\figurename}{Figure}
	\centering
	\begin{minipage}{1.0\linewidth}
		\centerline{\includegraphics[width=\textwidth]{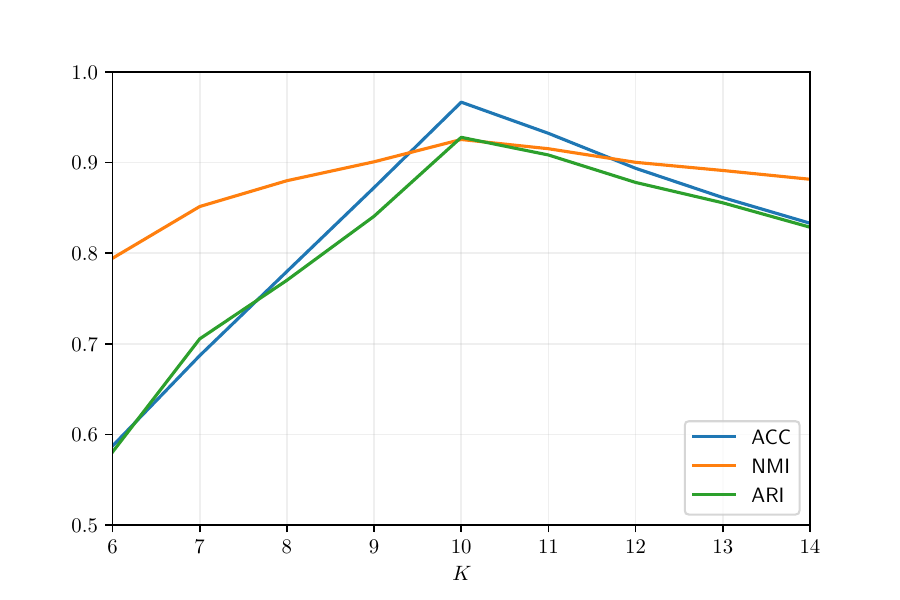}}
 	\end{minipage}
	\caption{Clustering performance v.s. parameter $K$.}
	\label{parameter analysis of K}
\end{figure}

\subsection{Ablation Study}\label{subsection:Ablation Study}
From the foregoing, we can see that the main contributions of the proposed DMSC are using the prior pairwise constraint information and introducing the (view-specific/common) feature properties protection paradigm to jointly carry out weighted multi-view representation learning and coherent clustering assignment. Therefore, this subsection focuses on exploring the importance of the semi-supervised (SEMI) module and the feature space protection (FSP) mechanism. Table \ref{ablation study 1c} and Table \ref{ablation study 3c} reveal the ablation results, where whether to fit out a specific part in DMSC is marked by ``$\checkmark$'' or ``$\times$,'' and from which something could be seen that when we individually add one of two parts to the benchmark model \cite{DMJCS}, enhanced performance can be observed in almost all cases. Furthermore, when both SEMI and FSP are simultaneously considered, our DMSC algorithm realizes the best clustering performance on the eight popular image datasets for all metrics. This observation legibly demonstrates that it is a very natural motivation to integrate semi-supervised learning paradigm and feature space preservation mechanism into deep multi-view clustering model, because the prior knowledge of pairwise constraints can better guide the intact clustering progress to obtain a more robust cluster-oriented shared representation based on the innovativeness of feature properties protection, shaping a perfect clustering structure and achieving an ideal clustering performance.

\subsection{Parameter Analysis}\label{subsection:Parameter Analysis}
In this subsection, we will discuss how four hyper-parameters, i.e., the clustering loss coefficient $\gamma$, the constraint loss coefficient $\lambda$, the prior knowledge proportion $\beta$ and the number of clusters $K$, affect the performance of the proposed DMSC.

We first probe into the sensitivity of $\gamma$, which is attached on the clustering term to protect feature properties. As exhibited from the comparative experiments in Section \ref{subsection:Experimental Comparison}, our DMSC works well with $\gamma = 0.1$. Figure \ref{parameter analysis of gamma} shows how our model performs with different $\gamma$ values. When $\gamma = 0$, the clustering constraint loses efficacy, leading to poor performance. When $\gamma$ raises gradually, the KL divergence based clustering constraint returns to life and enhanced clustering performance is acquired. In addition, with the increasement of $\gamma$, the fluctuation of metrics is considerably mild, which means that our model yields satisfactory performance for a suitable range of $\gamma$ and demonstrates that the proposed method is desensitized to the specific value of $\gamma$.

Next, the susceptibility variation of $\lambda$ is presented in Figure \ref{parameter analysis of lambda}, from which the three metrics' values soar as $\lambda$ changes from zero to non-zero, then they maintain perfect stability as $\lambda$ appropriately rises. This observation clearly suggests that when we consider a small amount of pairwise constraints prior knowledge, the model can make good use of this kind of valuable information to provoke a preferable representation learning capability and a superior clustering expressiveness.

After that, we analyze the parameter $\beta$ that renders $\beta \times n$ paired-sample constraints (or in other words $2 \times \beta \times n$ one-sample constraints) for the model training. As seen from Figure \ref{parameter analysis of beta}, as $\beta$ increases, the performance of DMSC generally promotes at the beginning and achieves stability in a wide range of $\beta$, which suggests that the incorporation of such pairwise constraint based semi-supervised learning rule into deep multi-view clustering model can result in a satisfactory performance via prior information capture.

With regard to the number of clusters $K$, we have assumed that $K$ on each dataset is predefined based on the ground-truth labels in the preceding experiments in Section \ref{subsection:Experimental Comparison} and Section \ref{subsection:Ablation Study}. Nevertheless, this is a strong assumption. In many real-world clustering applications, $K$ is usually unknown. Hence, we run our model on the STL10 dataset with different $K$ to search for the optimal value. As presented in Figure \ref{parameter analysis of K}, we can see that our model achieves the highest scores when $K = 10$, i.e., our model tends to group these objects into ten clusters, which is in accordance with the ground-truth labels.

\begin{table}
	\renewcommand{\arraystretch}{1.25}
	\caption{The robustness study for the number of views.}
	\label{robustness study table}
	\centering
	\scalebox{0.8}{
		\begin{threeparttable}
			\begin{tabular}{l|l|ccc}
				\hline
				Stage			&Method			&ACC&NMI&ARI\\
				\hline
				Initialization	&View1			&0.7112&0.6926&0.5940\\
				{}				&View2			&0.7317&0.7169&0.6348\\
				{}				&View3			&0.7252&0.7114&0.6268\\
				{}				&View1,2,3		&0.7492&0.7458&0.6678\\
				\hline
				Finetuning		&View1,2			&0.7866&0.8163&0.7380\\
				{}				&View1,3			&0.7782&0.8131&0.7362\\
				{}				&View2,3			&0.7804&0.8119&0.7317\\
				{}				&View1,2,3		&\textbf{0.7873}&\textbf{0.8263}&\textbf{0.7452}\\
				\hline
			\end{tabular}
			\footnotesize
	\end{threeparttable}}
\end{table}

\begin{figure*}
	\setlength{\abovecaptionskip}{0pt}
	\setlength{\belowcaptionskip}{0pt}
	\renewcommand{\figurename}{Figure}
	\centering
	\begin{minipage}{0.25\linewidth}
		\centerline{\includegraphics[width=\textwidth]{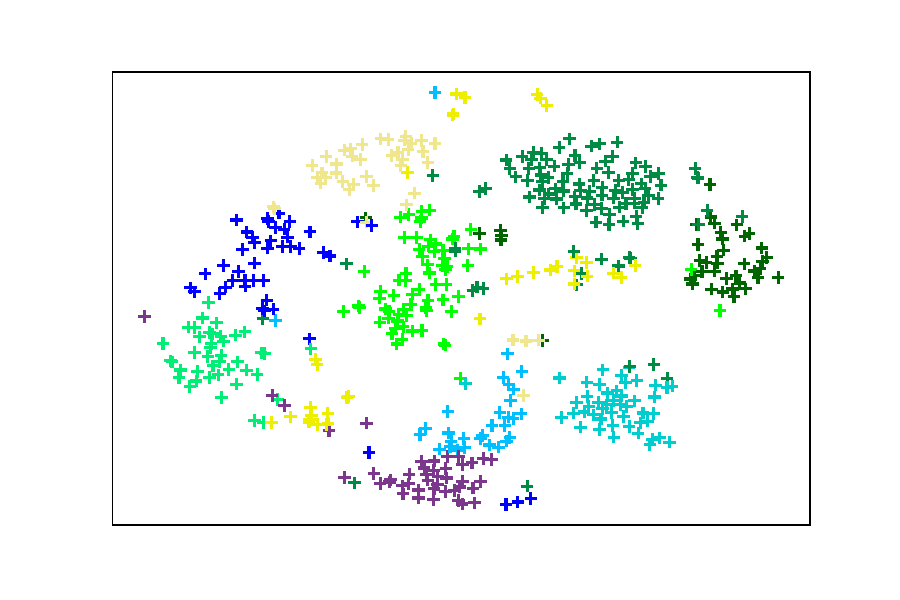}}
		\vspace{-10pt}
		\centerline{(a) View1}
		\vspace{5pt}
	\end{minipage}
	\hspace{-15pt}
	\begin{minipage}{0.25\linewidth}
		\centerline{\includegraphics[width=\textwidth]{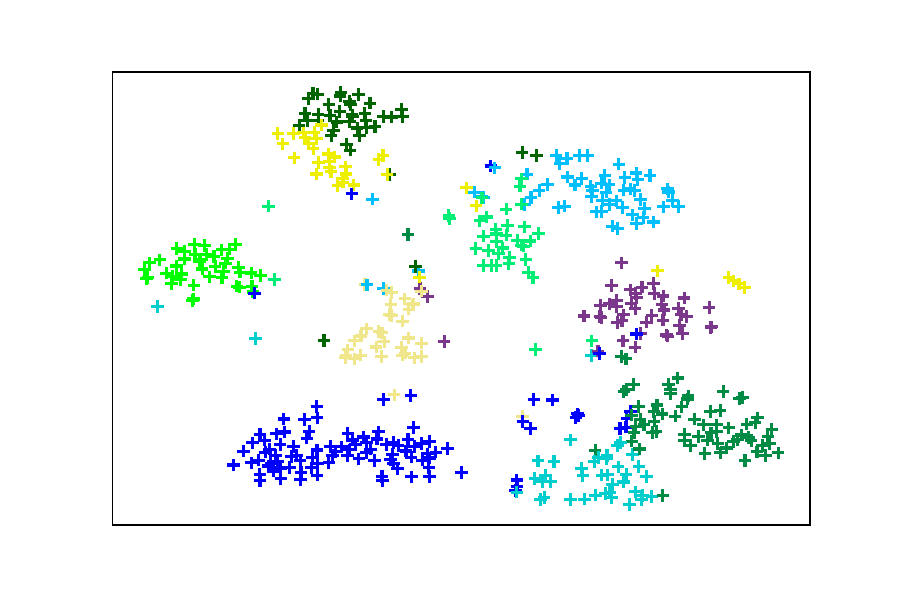}}
		\vspace{-10pt}
		\centerline{(b) View2}
		\vspace{5pt}
	\end{minipage}
	\hspace{-15pt}
	\begin{minipage}{0.25\linewidth}
		\centerline{\includegraphics[width=\textwidth]{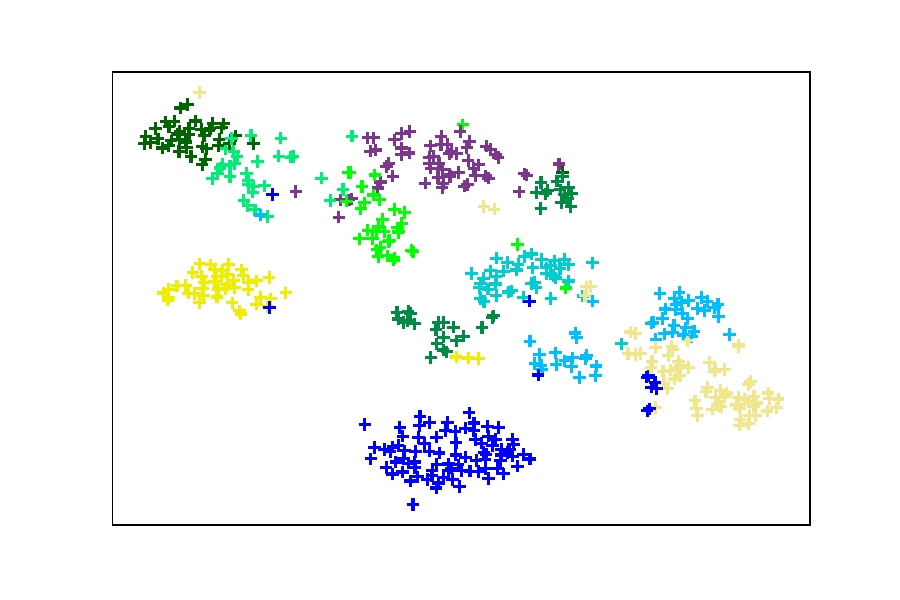}}
		\vspace{-10pt}
		\centerline{(c) View3}
		\vspace{5pt}
	\end{minipage}
	\hspace{-15pt}
	\begin{minipage}{0.25\linewidth}
		\centerline{\includegraphics[width=\textwidth]{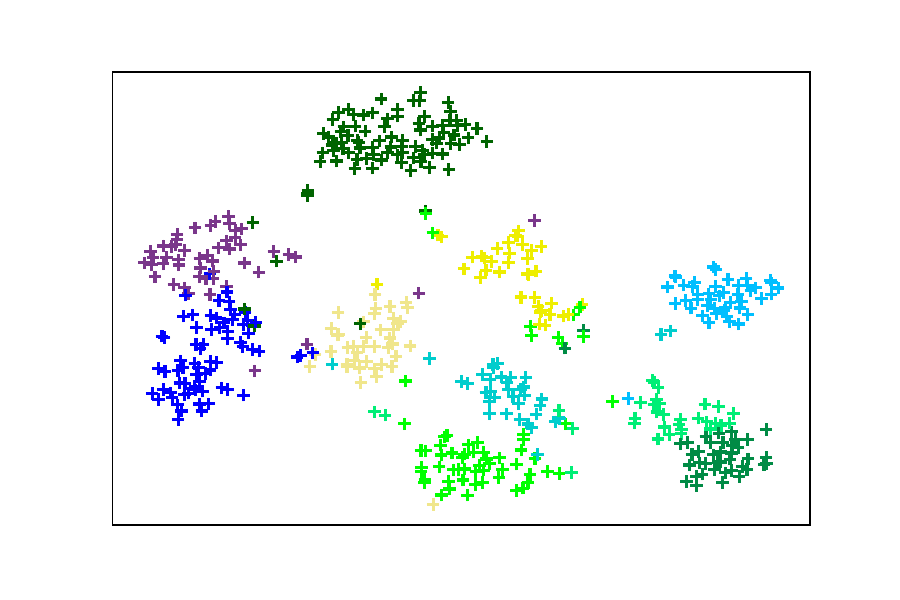}}
		\vspace{-10pt}
		\centerline{(d) View1,2,3}
		\vspace{5pt}
	\end{minipage}
	\caption{$t$-SNE visualization of the clusters after parameters initialization.}
	\label{robustness study tsne}
\end{figure*}

\subsection{Robustness Study}
Note that in the previous experiments in Section \ref{subsection:Experimental Comparison}, Section \ref{subsection:Ablation Study}, Section \ref{subsection:Parameter Analysis}, we have assumed that the number of views $V$ on each dataset is given as two. However, in many real-world applications, $V$ is usually greater than that. Consequently, here we run our model on the USPS dataset with three multiple deep branches ($V = 3$) equipped to study the robustness with regard to the view numbers, whose quality is also measured by ACC/NMI/ARI. To be specific, the SAE and CAE defined in Table \ref{exp config} are considered as the first two views, and a variational autoencoder (VAE) is utilized as the third view. Its encoding architecture is $\text{Conv}_{6}^{2}-\text{Conv}_{20}^{3}-\text{Conv}_{60}^{3}-\text{Fc}_{256}-\text{Fc}_{10}$, where $\text{Conv}_{n}^{k}$ refers to a convolutional layer with $n$ filters, $k \times k$ kernel size and $2$ stride length, $\text{Fc}_{n}$ represents a fully connected layer with $n$ neurons. Naturally, the mirrored version of the encoder is deemed as the decoding network. The results of the robustness experiment are presented in Table \ref{robustness study table} and Figure \ref{robustness study tsne}. Generally speaking, simply concatenating three heterogeneous features does bring better initialization performance than single view, see the numerical comparisons of View1,2,3 (multi-view concatenated feature) v.s. View1/2/3 (single-view feature) in \emph{Stage-Initialization} from Table \ref{robustness study table} and the $t$-SNE \cite{t-SNE} visualization from Figure \ref{robustness study tsne}. Meanwhile, as finetuning iteratively proceeds until model convergence, the proposed approach achieves more brilliant clustering performance than two views ones, which implies that three types of feature embeddings obtained by SAE, CAE, VAE can nicely complement each other and boost the clustering uniformity under our DMSC framework, see the clustering results in \emph{Stage-Finetuning} from Table \ref{robustness study table}. In one word, the DMSC method owns a relatively good generalization for the number of views $V$.

\begin{figure*}
	\setlength{\abovecaptionskip}{0pt}
	\setlength{\belowcaptionskip}{0pt}
	\renewcommand{\figurename}{Figure}
	\centering
	\begin{minipage}{0.25\linewidth}
		\centerline{\includegraphics[width=\textwidth]{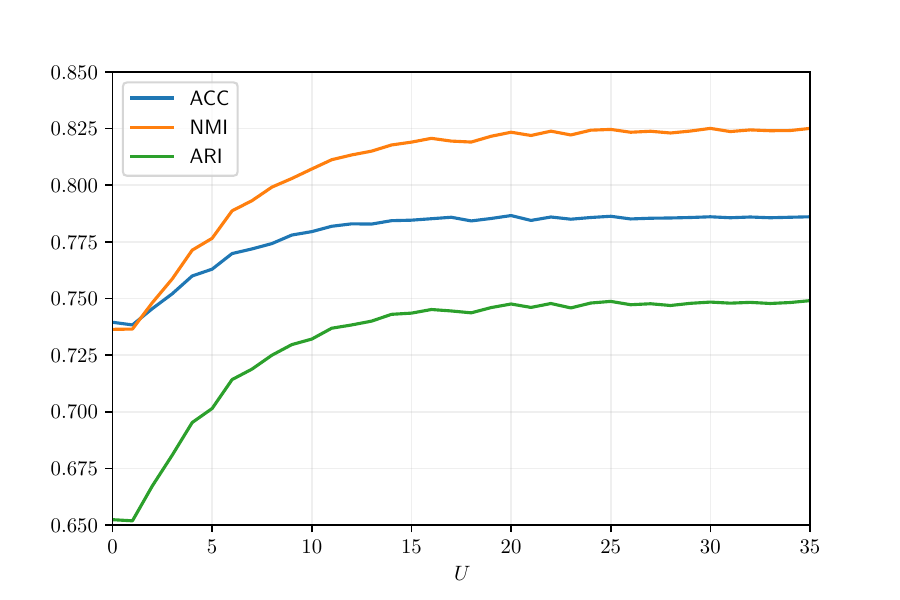}}
		\centerline{(a) USPS}
		\vspace{5pt}
	\end{minipage}
	\hspace{-15pt}
	\begin{minipage}{0.25\linewidth}
		\centerline{\includegraphics[width=\textwidth]{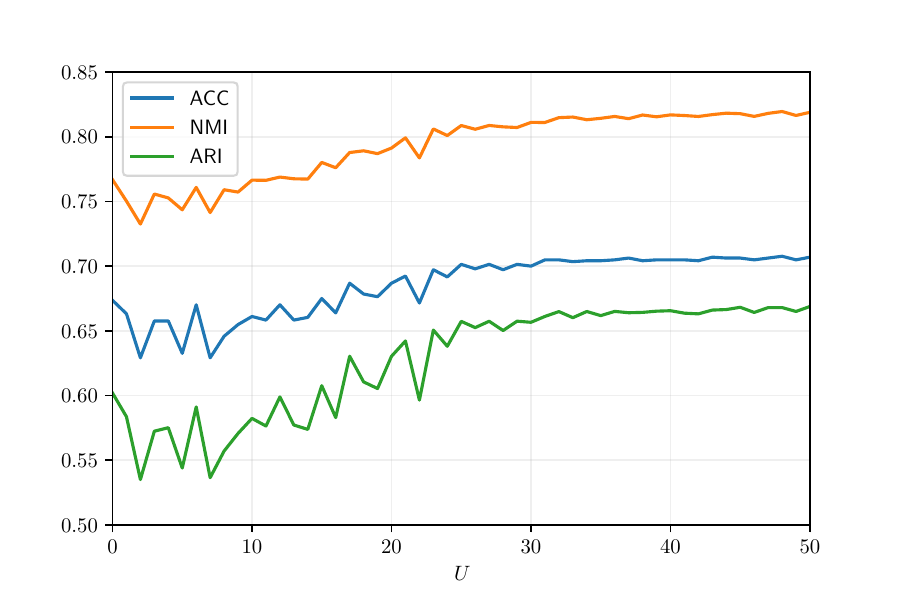}}
		\centerline{(b) COIL20}
		\vspace{5pt}
	\end{minipage}
	\hspace{-15pt}
	\begin{minipage}{0.25\linewidth}
		\centerline{\includegraphics[width=\textwidth]{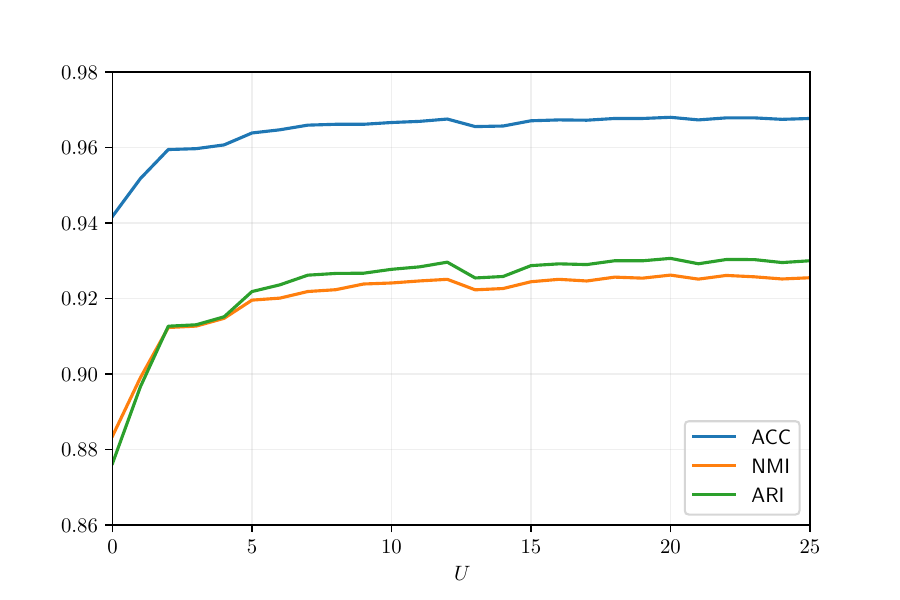}}
		\centerline{(c) STL10}
		\vspace{5pt}
	\end{minipage}
	\hspace{-15pt}
	\begin{minipage}{0.25\linewidth}
		\centerline{\includegraphics[width=\textwidth]{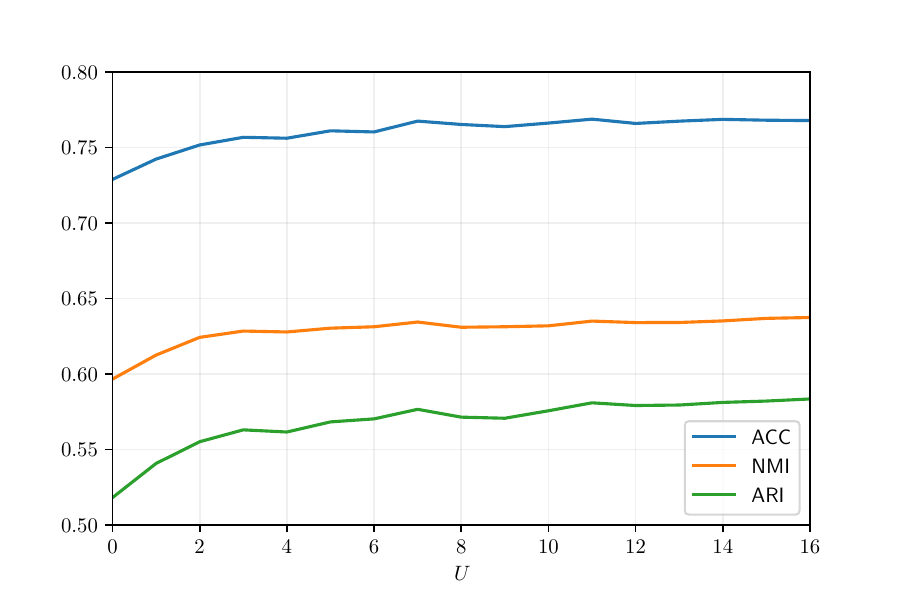}}
		\centerline{(d) CIFAR10}
		\vspace{5pt}
	\end{minipage}
	\caption{Clustering performance v.s. iterations.}
	\label{convergence analysis metrics}
\end{figure*}

\begin{figure*}
	\setlength{\abovecaptionskip}{0pt}
	\setlength{\belowcaptionskip}{0pt}
	\renewcommand{\figurename}{Figure}
	\centering
	\begin{minipage}{0.25\linewidth}
		\centerline{\includegraphics[width=\textwidth]{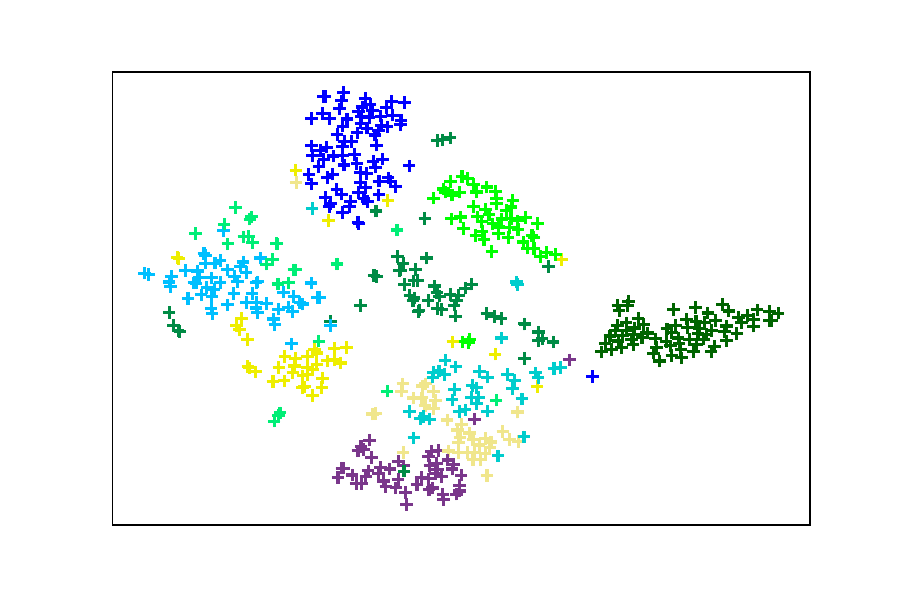}}
		\vspace{-10pt}
		\centerline{(a) Original}
		\vspace{5pt}
	\end{minipage}
	\hspace{-15pt}
	\begin{minipage}{0.25\linewidth}
		\centerline{\includegraphics[width=\textwidth]{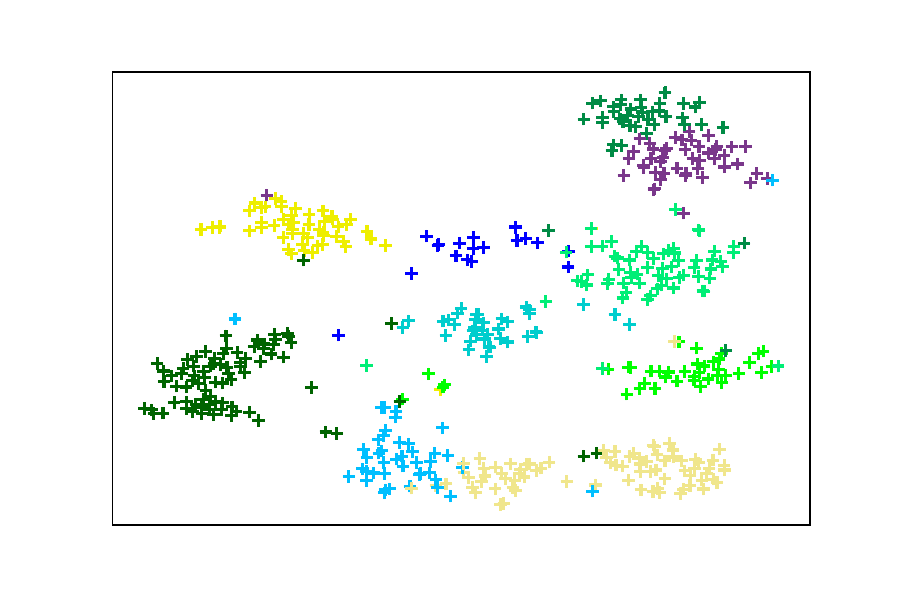}}
		\vspace{-10pt}
		\centerline{(b) Initial}
		\vspace{5pt}
	\end{minipage}
	\hspace{-15pt}
	\begin{minipage}{0.25\linewidth}
		\centerline{\includegraphics[width=\textwidth]{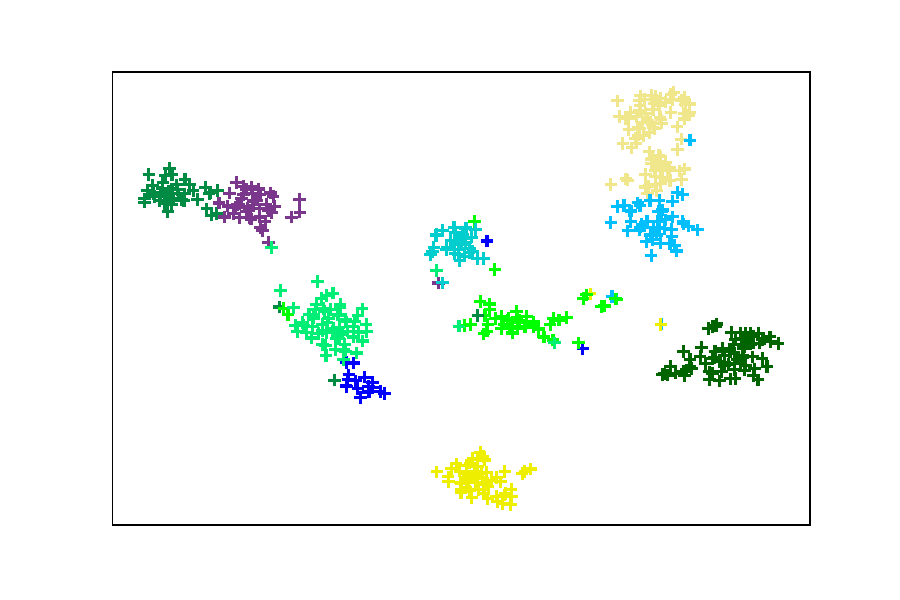}}
		\vspace{-10pt}
		\centerline{(c) Iterative}
		\vspace{5pt}
	\end{minipage}
	\hspace{-15pt}
	\begin{minipage}{0.25\linewidth}
		\centerline{\includegraphics[width=\textwidth]{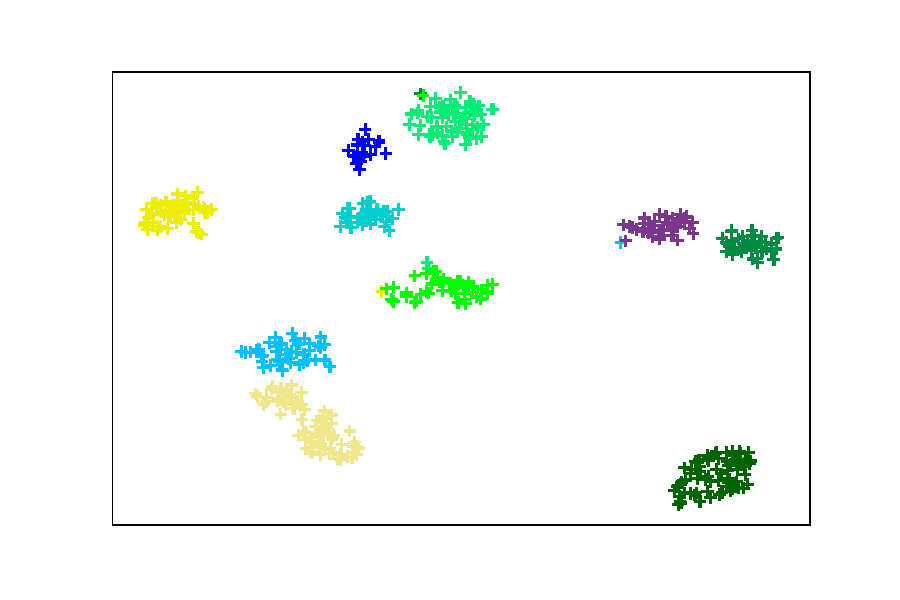}}
		\vspace{-10pt}
		\centerline{(d) Final}
		\vspace{5pt}
	\end{minipage}
	\caption{$t$-SNE visualization of the clusters during networks finetuning.}
	\label{convergence analysis tsne}
\end{figure*}

\subsection{Convergence Analysis}
To study the convergence of the proposed DMSC, we first record the three evaluation metrics over iterations on the datasets USPS, COIL20, STL10, CIFAR10. As can be observed from the results described in Figure \ref{convergence analysis metrics}, there is a distinct upward trend of each metric in the first few iterations, and all metrics eventually reach stability. Moreover, we use $t$-SNE \cite{t-SNE} to visualize the learned common representation in different periods of the training process on a subset of the USPS dataset with $500$ samples, see Figure \ref{convergence analysis tsne}. We can heed that feature points mapped from raw pixels are extremely overlapped, implying the challenge of the clustering task, as Figure \ref{convergence analysis tsne}-(a). After parameters initialization shown in Figure \ref{convergence analysis tsne}-(b), the distribution of the merged features embedded by multiple deep branches is more discrete than the original features, and the preliminary clustering structure has been formed, this is because the learned initial representations are high-level and cluster-oriented. As finetuning proceeds until the model achieves convergence, feature points remain steady and are nicely separated, as Figure \ref{convergence analysis tsne}-(c)(d) displayed. Overall, Figure \ref{convergence analysis metrics} and Figure \ref{convergence analysis tsne} indeed illustrate that our DMSC can converge practically.

\section{Conclusion}\label{section:Conclusion}
In this paper, we propose a novel deep multi-view semi-supervised clustering approach DMSC, which can boost the performance of multi-view clustering effectively by deriving the weakly-supervised information contained in sample pairwise constraints and protecting the feature properties of multi-view data. Using pairwise constraints prior knowledge during model training is beneficial for shaping a reliable clustering structure. The feature properties protection mechanism effectively prevents view-specific and view-shared feature from being distorted in clustering optimization. In comparison with existing state-of-the-art single-view and multi-view clustering competitors, the proposed method achieves the best performance on eight benchmark image datasets. Future work may cover conducting more trials on large-scale image, text, audio datasets with multiple views to ensure the model generalization and further exploring more advanced multi-view weighting technique for robust common representation learning to enhance multi-view clustering performance.

\section*{Acknowledgment}
The authors are thankful for the financial support by the National Key Research and Development Program of China (no. 2020AAA0109500), the Key-Area Research and Development Program of Guangdong Province (no. 2019B010153002), the National Natural Science Foundation of China (nos. 62106266, 61961160707, 61976212, U1936206, 62006139).


\begin{thebibliography}{99}

\bibitem{KM}
J. MacQueen, ``Some methods for classification and analysis of multivariate observations,'' in \emph{Berkeley Symposium on Mathematical Statistics and Probability}, pp. 281-297, 1967.
\bibitem{DEC}
J. Xie, R. Girshick, and A. Farhadi, ``Unsupervised deep embedding for clustering analysis,'' in \emph{International Conference on Machine Learning}, pp. 478-487, 2016.
\bibitem{SC}
J. Shi and J. Malik, ``Normalized cuts and image segmentation,'' \emph{IEEE Transactions on Pattern Analysis and Machine Intelligence}, vol. 22, no. 8, pp. 888-905, 2000.
\bibitem{SDCN}
D. Bo, X. Wang, C. Shi, M. Zhu, E. Lu, and P. Cui, ``Structural deep clustering network,'' in \emph{The Web Conference}, pp. 1400-1410, 2020.
\bibitem{GMM}
C. M. Bishop, ``Pattern recognition and machine learning,'' Springer, 2006.
\bibitem{DGG}
L. Yang, N. Cheung, J. Li, and J. Fang, ``Deep clustering by gaussian mixture variational autoencoders with graph embedding,'' in \emph{IEEE International Conference on Computer Vision}, pp. 6439-6448, 2019.
\bibitem{DBSCAN}
M. Ester, H. P. Kriegel, J. Sander, and X. Xu, ``A density-based algorithm for discovering clusters in large spatial databases with noise,'' in \emph{ACM SIGKDD Conference on Knowledge Discovery and Data Mining}, pp. 226-231, 1996.
\bibitem{DDC}
Y. Ren, N. Wang, M. Li, and Z. Xu, ``Deep density-based image clustering,'' \emph{Knowledge-Based Systems}, vol. 197, pp. 105841, 2020.
\bibitem{Gabor}
M. Lades, J. C. Vorbruggen, J. Buhmann, J. Lange, C. V. D. Malsburg, R. P. Wurtz, and W. Konen, ``Distortion invariant object recognition in the dynamic link architecture,'' \emph{IEEE Transactions on Computers}, vol. 42, no. 3, pp. 300-311, 1993.
\bibitem{LBP}
T. Ojala, M. Pietikainen, and T. Maenpaa, ``Multiresolution gray-scale and rotation invariant texture classification with local binary patterns,'' \emph{IEEE Transactions on Pattern Analysis and Machine Intelligence}, vol. 24, no. 7, pp. 971-987, 2002.
\bibitem{SIFT}
D. G. Lowe, ``Distinctive image features from scale-invariant keypoints,'' \emph{International Journal of Computer Vision}, vol. 60, no. 2, pp. 91-110, 2004.
\bibitem{HOG}
N. Dalal and B. Triggs, ``Histograms of oriented gradients for human detection,'' in \emph{IEEE Conference on Computer Vision and Pattern Recognition}, pp. 886-893, 2005.
\bibitem{CCA}
H. Hotelling, ``Relations between two sets of variates,'' \emph{Biometrika}, vol. 28, pp. 321-377, 1936.
\bibitem{KCCA}
S. Akaho, ``A kernel method for canonical correlation analysis,'' \emph{arXiv preprint arXiv:cs/0609071}, 2006.
\bibitem{MVSC}
H. Gao, F. Nie, X. Li, and H. Huang, ``Multi-view subspace clustering,'' \emph{IEEE International Conference on Computer Vision}, pp. 4238-4246, 2015.
\bibitem{LMSC}
C. Zhang, H. Fu, Q. Hu, X. Cao, Y. Xie, D. Tao, and D. Xu, ``Generalized latent multi-view subspace clustering,'' \emph{IEEE Transactions on Pattern Analysis and Machine Intelligence}, vol. 42, no. 1, pp. 86-99, 2020.
\bibitem{CoMSC}
J. Liu, X. Liu, Y. Yang, X. Guo, M. Kloft, and L. He, ``Multiview subspace clustering via co-training robust data representation,'' \emph{IEEE Transactions on Neural Networks and Learning Systems}, 2021, doi: 10.1109/TNNLS.2021.3069424.
\bibitem{CTRL}
Y. Tang, Y. Xie, C. Zhang, and W. Zhang, ``Constrained tensor representation learning for multi-view semi-supervised subspace clustering,'' \emph{IEEE Transactions on Multimedia}, 2021, doi: 10.1109/TMM.2021.3110098.
\bibitem{JSTC}
Y. Tang, Y. Xie, C. Zhang, Z. Zhang, and W. Zhang, ``One-step multiview subspace segmentation via joint skinny tensor learning and latent clustering,'' \emph{IEEE Transactions on Cybernetics}, 2021, doi: 10.1109/TCYB.2021.3053057.
\bibitem{T-MEK-SPL}
Y. Tang, Y. Xie, X. Yang, J. Niu, and W. Zhang, ``Tensor multi-elastic kernel self-paced learning for time series clustering,'' \emph{IEEE Transactions on Knowledge and Data Engineering}, vol. 33, no. 3, pp. 1223-1237, 2021.
\bibitem{DCCA}
G. Andrew, R. Arora, J. Bilmes, and K. Livescu, ``Deep canonical correlation analysis,'' in \emph{International Conference on Machine Learning}, pp. 1247-1255, 2013.
\bibitem{DCCAE}
W. Wang, R. Arora, K. Livescu, and J. Bilmes, ``On deep multi-view representation learning,'' in \emph{International Conference on Machine Learning}, pp. 1083-1092, 2015.
\bibitem{DGCCA}
A. Benton, H. Khayrallah, B. Gujral, D. Reisinger, S. Zhang, and R. Arora, ``Deep generalized canonical correlation analysis,'' \emph{arXiv preprint arXiv:1702.02519}, 2017.
\bibitem{DMJCS}
Y. Xie, B. Lin, Y. Qu, C. Li, W. Zhang, L. Ma, Y. Wen, and D. Tao, ``Joint deep multi-view learning for image clustering,'' \emph{IEEE Transactions on Knowledge and Data Engineering}, vol. 33, no. 11, pp. 3594-3606, 2021.
\bibitem{DEMVC}
J. Xu, Y. Ren, G. Li, L. Pan, C. Zhu, and Z. Xu ``Deep embedded multi-view clustering with collaborative training,'' \emph{Information Sciences}, vol. 573, pp. 279-290, 2021.
\bibitem{AE2-Nets}
C. Zhang, Y. Liu, and H. Fu, ``AE2-Nets: Autoencoder in autoencoder networks,'' in \emph{IEEE Conference on Computer Vision and Pattern Recognition}, pp. 2572-2580, 2019.
\bibitem{CDIMC-net}
J. Wen, Z. Zhang, Y. Xu, B. Zhang, L. Fei, and G. Xie, ``CDIMC-net: Cognitive deep incomplete multi-view clustering network,'' in \emph{International Joint Conference on Artificial Intelligence}, pp. 3230-3236, 2020.
\bibitem{semi-KM 1}
S. Basu, A. Banerjee, and R. J. Mooney, ``Active semi-supervision for pairwise constrained clustering,'' in \emph{IEEE International Conference on Data Mining}, pp. 333-344, 2004.
\bibitem{semi-KM 2}
P. Bradley, K. Bennett, and A. Demiriz, ``Constrained k-means clustering,'' \emph{Microsoft Research}, 2000.
\bibitem{semi-SC}
S. D. Kamvar, D. Klein, and C. D. Manning, ``Spectral learning,'' in \emph{International Joint Conference on Artificial Intelligence}, pp. 561-566, 2003.
\bibitem{DAC}
J. Chang, L. Wang, G. Meng, S. Xiang, and C. Pan, ``Deep adaptive image clustering,'' in \emph{IEEE International Conference on Computer Vision}, pp. 5880-5888, 2017.
\bibitem{ConPaC}
Y. Shi, C. Otto, and A. K. Jain, ``Face clustering: Representation and pairwise constraints,'' \emph{IEEE Transactions on Information Forensics and Security}, vol. 13, no. 7, pp. 1626-1640, 2018.
\bibitem{SSFPC}
Z. Wang, S. Wang, L. Bai, W. Wang, and Y. Shao, ``Semisupervised fuzzy clustering with fuzzy pairwise constraints,'' \emph{IEEE Transactions on Fuzzy Systems}, 2021, doi: 10.1109/TFUZZ.2021.3129848.
\bibitem{MLAN}
F. Nie, G. Cai, J. Li, and X. Li, ``Auto-weighted multi-view learning for image clustering and semi-supervised classification,'' \emph{IEEE Transactions on Image Processing}, vol. 27, no. 3, pp. 1501-1511, 2018.
\bibitem{SSSL-M}
Y. Qin, H. Wu, X. Zhang, and G. Feng, ``Semi-supervised structured subspace learning for multi-view clustering,'' \emph{IEEE Transactions on Image Processing}, vol. 31, pp. 1-14, 2022.
\bibitem{SC-MPI}
L. Bai, J. Liang, and F. Cao, ``Semi-supervised clustering with constraints of different types from multiple information sources,'' \emph{IEEE Transactions on Pattern Analysis and Machine Intelligence}, vol. 43, no. 9, pp. 3247-3258, 2021.
\bibitem{for readers 1}
S. Basu, I. Davidson, and K. Wagstaff, ``Constrained clustering: Advances in algorithms, theory, and applications,'' Chapman \& Hall/CRC, 2008.
\bibitem{for readers 2}
E. Bair, ``Semi-supervised clustering methods,'' \emph{Wiley Interdisciplinary Reviews Computational Statistics}, vol. 5, no. 5, pp. 349-361, 2013.
\bibitem{review}
X. Yan, S. Hu, Y. Mao, Y. Ye, and H. Yu, ``Deep multi-view learning methods: A review,'' \emph{Neurocomputing}, vol. 448, pp. 106-129, 2021.
\bibitem{SAMVC}
Y. Ren, S. Huang, P. Zhao, M. Han, and Z. Xu, ``Self-paced and auto-weighted multi-view clustering,'' \emph{Neurocomputing}, vol. 383, pp. 248-256, 2020.
\bibitem{IDEC}
X. Guo, L. Gao, X. Liu, and J. Yin, ``Improved deep embedded clustering with local structure preservation,'' in \emph{International Joint Conference on Artificial Intelligence}, pp. 1753-1759, 2017.
\bibitem{DCN}
B. Yang, X. Fu, N. D. Sidiropoulos, and M. Hong, ``Towards kmeans-friendly spaces: Simultaneous deep learning and clustering,'' in \emph{International Conference on Machine Learning}, pp. 3861-3870, 2017.
\bibitem{ASPC}
X. Guo, X. Liu, E. Zhu, X. Zhu, M. Li, X. Xu, and J. Yin, ``Adaptive self-paced deep clustering with data augmentation,'' \emph{IEEE Transactions on Knowledge and Data Engineering}, vol. 32, no. 9, pp. 1680-1693, 2020.
\bibitem{SDEC}
Y. Ren, K. Hu, X. Dai, L. Pan, S. C. H. Hoi, and Z. Xu, ``Semi-supervised deep embedded clustering,'' \emph{Neurocomputing}, vol. 325, pp. 121-130, 2019.
\bibitem{DBC}
F. Li, H. Qiao, and B. Zhang, ``Discriminatively boosted image clustering with fully convolutional auto-encoders,'' \emph{Pattern Recognition}, vol. 83, pp. 161-173, 2018.
\bibitem{DKM}
M. M. Fard, T. Thonet, and E. Gaussier, ``Deep k-means: Jointly clustering with k-means and learning representations,'' \emph{Pattern Recognition Letters}, vol. 138, pp. 185-192, 2020.
\bibitem{DCSPC}
R. Chen, Y. Tang, L. Tian, C. Zhang, and W. Zhang, ``Deep convolutional self-paced clustering,'' \emph{Applied Intelligence}, 2021, doi: 10.1007/s10489-021-02569-y.
\bibitem{DMNEC}
R. Chen, Y. Tang, C. Zhang, W. Zhang, and Z. Hao, ``Deep multi-network embedded clustering,'' \emph{Pattern Recognition and Artificial Intelligence}, vol. 34, no. 1, pp. 14-24, 2021.
\bibitem{DCMSC}
Z. Li, C. Tang, J. Chen, C. Wan, W. Yan, and X. Liu, ``Diversity and consistency learning guided spectral embedding for multi-view clustering,'' \emph{Neurocomputing}, vol. 370, pp. 128-139, 2019.
\bibitem{MCIM}
D. Wu, Z. Hu, F. Nie, R. Wang, H. Yang, and X. Li, ``Multi-view clustering with interactive mechanism,'' \emph{Neurocomputing}, vol. 449, pp. 378-388, 2021.
\bibitem{RMKMC}
X. Cai, F. Nie, and H. Huang, ``Multi-view k-means clustering on big data,'' in \emph{International Joint Conference on Artificial Intelligence}, pp. 2598-2604, 2013.
\bibitem{MSPL}
C. Xu, D. Tao, and C. Xu, ``Multi-view self-paced learning for clustering,'' in \emph{International Joint Conference on Artificial Intelligence}, pp. 3974-3980, 2015.
\bibitem{BMVC}
Z. Zhang, L. Liu, F. Shen, H. T. Shen, and L. Shao, ``Binary multi-view clustering,'' \emph{IEEE Transactions on Pattern Analysis and Machine Intelligence}, vol. 41, no. 7, pp. 1774-1782, 2019.
\bibitem{ACC}
T. Li and C. Ding, ``The relationships among various nonnegative matrix factorization methods for clustering,'' in \emph{IEEE International Conference on Data Mining}, pp. 362-371, 2006.
\bibitem{NMI}
A. Strehl and J. Ghosh, ``Cluster ensembles: A knowledge reuse framework for combining multiple partitions,'' \emph{Journal of Machine Learning Research}, vol. 3, pp. 583-617, 2002.
\bibitem{ARI}
L. Hubert and P. Arabie, ``Comparing partitions,'' \emph{Journal of Classification}, vol. 2, no. 1, pp. 193-218, 1985.
\bibitem{Hungarian algorithm}
H. W. Kuhn, ``The hungarian method for the assignment problem,'' \emph{Naval Research Logistics Quarterly}, vol. 2, no. 1, pp. 83-97, 1955.
\bibitem{RI}
W. M. Rand, ``Objective criteria for the evaluation of clustering methods,'' \emph{Journal of the American Statistical Association}, vol. 66, no. 336, pp. 846-850, 1971.
\bibitem{Adam}
D. Kingma and J. Ba, ``Adam: A method for stochastic optimization,'' \emph{arXiv preprint arXiv:1412.6980}, 2014.
\bibitem{ReLU}
X. Glorot, A. Bordes, and Y. Bengio, ``Deep sparse rectifier neural networks,'' \emph{Journal of Machine Learning Research}, vol. 15, pp. 315-323, 2011.
\bibitem{Xariver}
X. Glorot and Y. Bengio, ``Understanding the difficulty of training deep feedforward neural networks,'' \emph{Journal of Machine Learning Research}, vol. 9, pp. 249-256, 2010.
\bibitem{t-SNE}
L. V. D. Maaten and G. Hinton, ``Visualizing data using t-sne,'' \emph{Journal of Machine Learning Research}, vol. 9, no. 12, pp. 2579-2605, 2008.

\end{thebibliography}
\end{document}